\documentclass[lettersize,journal]{IEEEtran}
\usepackage{amsmath,amsfonts}
\usepackage{algorithm}
\usepackage{algorithmic}
\usepackage{array}
\usepackage[caption=false,font=normalsize,labelfont=sf,textfont=sf]{subfig}
\usepackage{textcomp}
\usepackage{stfloats}
\usepackage{xcolor}         % colors
\usepackage{url}
\usepackage{verbatim}
\usepackage{graphicx}
\usepackage{amssymb}
\usepackage{mathtools}
\usepackage{amsthm}
\usepackage{booktabs}
\usepackage{multirow}
\usepackage{cite}
\usepackage{enumitem}
\usepackage[table]{xcolor}
\usepackage{hyperref}       % hyperlinks
\usepackage{fontawesome5}

\newcommand{\lsy}[1]{\textcolor{black}{#1}}
\definecolor{ours}{RGB}{255,245,235} % 淡橙
\definecolor{lightgray}{RGB}{245,245,255}
\definecolor{lightgreen}{RGB}{235,250,235}
\definecolor{reda}{RGB}{192,0,0}
\hypersetup{
	colorlinks   = true,
	urlcolor     = blue,
	linkcolor    = reda,
	citecolor   = blue
}

\definecolor{scopeRow}{HTML}{F4F0FF}   % very light purple: full method
\definecolor{deltaPos}{HTML}{2E7D32}   % muted green: improvement
\definecolor{deltaNeg}{HTML}{C62828}   % muted red: degradation
\definecolor{deltaZero}{HTML}{666666}  % neutral gray: unchanged

\newcommand{\dpos}[1]{{\scriptsize\textcolor{deltaPos}{\,($\uparrow #1$)}}}
\newcommand{\dneg}[1]{{\scriptsize\textcolor{deltaNeg}{\,($\downarrow #1$)}}}
\newcommand{\dzero}{{\scriptsize\textcolor{deltaZero}{\,($\pm0.00$)}}}

\begin{document}

\title{Multimodal Knowledge Edit-Scoped Generalization for Online Recursive  MLLM Editing}

\author{Siyuan Li, Youyuan Zhang, Ruitong Liu, Junxi Wang and Jing Li \textsuperscript{\faEnvelope[regular]},~\IEEEmembership{Senior Member,~IEEE} 
\thanks{\faEnvelope[regular] Jing Li is the corresponding author (E-mail:
jingli.phd@hotmail.com).}
\thanks{Siyuan Li and Youyuan Zhang are with Harbin Institute of Technology, Shenzhen, China and Peng Cheng Laboratory. Ruitong Liu is with Peking University. Junxi Wang is with Fudan University. Jing Li is with Harbin Institute of Technology, Shenzhen, China.}}

% The paper headers
\markboth{}%
{Li \MakeLowercase{\textit{et al.}}: Multimodal Knowledge Edit-Scoped Generalization for Online Recursive  MLLM Editing}

\maketitle

\begin{abstract}
Online multimodal knowledge editing requires injecting a continual stream of visual-textual corrections into multimodal large language models (MLLMs) with bounded overhead and minimal disruption to unrelated behaviors. 
Existing editors mainly emphasize edit reliability and long-horizon stability, but rarely control the semantic boundary of each edit. 
Our pilot analyses of post-edit behaviors and internal neuronal activities reveal a scope gap behind reliable edits: instance-level success neither guarantees transfer to valid cross-modal variants nor prevents leakage to unrelated inputs, while edit-related cross-modal responses concentrate in deeper semantic layers. 
Therefore, we formulate \emph{Edit-Scoped Generalization}, reframing online MLLM editing from merely correcting an instance to controlling the propagation boundary of each edit. 
To this end, we propose \textbf{ScopeEdit}, a scope-aware online editor that decomposes each update into a modality-local absorption branch and an evidence-gated shared generalization branch. 
The local branch supports stable edit absorption, whereas the shared branch enables cross-modal propagation only when visual and textual evidence are sufficiently aligned. 
Both branches perform scope-separated write geometries in orthogonal low-rank spaces and maintain branch-wise preconditioners via Sherman--Morrison recursions, yielding constant per-edit overhead.
Extensive experiments across diverse benchmarks, long-horizon edit streams, MLLM backbones, real-world VLKEB scenarios, and complex vision-language architectures show that ScopeEdit consistently improves the trade-off between in-scope cross-modal transfer and out-of-scope locality, while preserving edit reliability, stability and online efficiency.
Our code is available at \url{https://github.com/lab-klc/ScopeEdit}.
\end{abstract}


\section{Introduction}
\begin{figure*} 
\centering 
\includegraphics[width=1\linewidth]{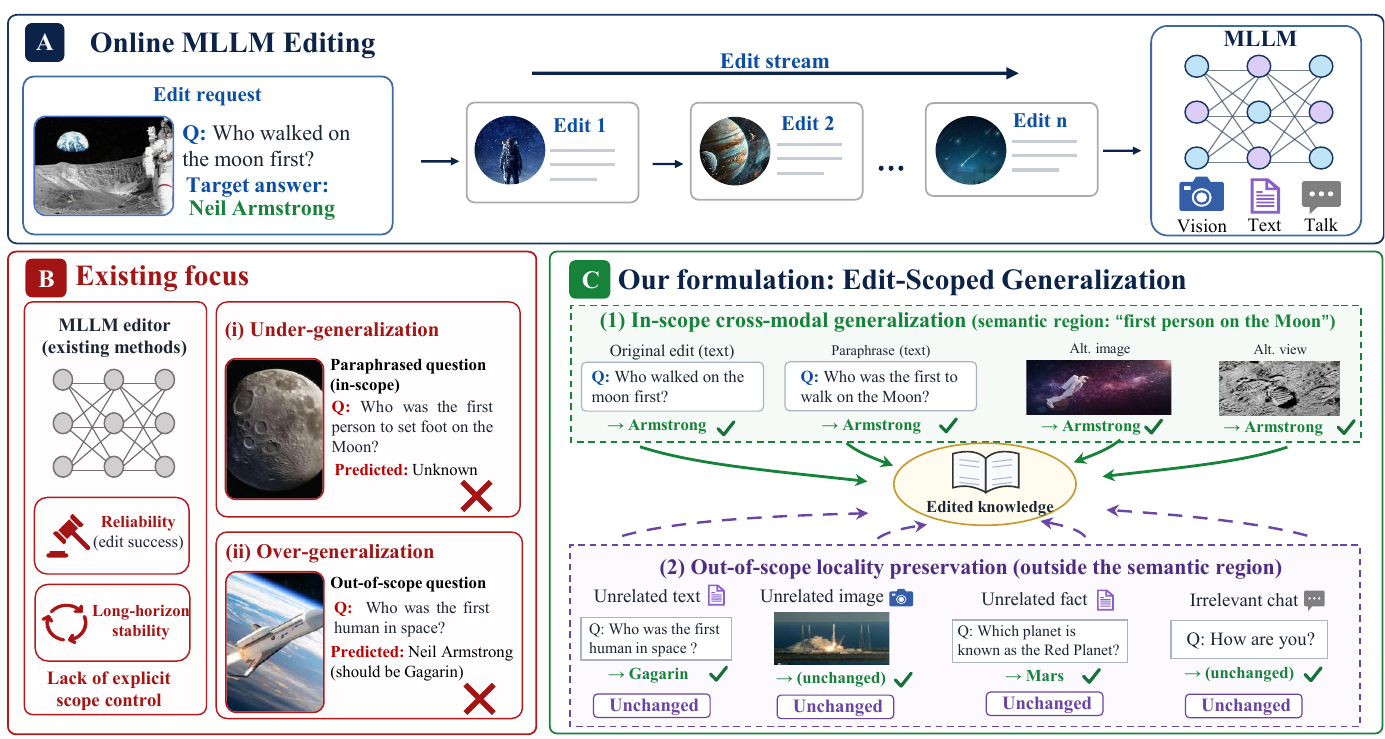} 
\caption{Motivation and formulation of \emph{edit-scoped generalization} in online MLLM editing. \textbf{(A)} Online multimodal knowledge editing incorporates a stream of visual-textual corrections into an MLLM. \textbf{(B)} Existing editors mainly focus on edit reliability and long-horizon stability, but lack explicit control over the semantic scope of each edit, leading to under-generalization on in-scope variants or over-generalization to unrelated inputs. \textbf{(C)} We define edit-scoped generalization as the joint requirement of in-scope cross-modal generalization and out-of-scope locality preservation.} 
\label{fig:introduction}
\vspace{-3mm}
\end{figure*}

\IEEEPARstart{M}{ultimodal} large language models (MLLMs) have become a dominant paradigm for knowledge-intensive multimodal tasks~\cite{touvron2023llama,zhao2023survey,li2023blip,lin2024video,lin2026moe,xue2026vision}. 
However, their visual and textual knowledge is encoded in static parameters, whose validity inevitably changes as real-world facts evolve~\cite{jiang2026when,wang2026vlbiasbench}. 
\emph{Online multimodal knowledge editing} aims to incorporate a stream of corrections into an MLLM without retraining from scratch, while preserving unrelated capabilities and behaviors~\cite{meng2022locating,fangalphaedit,chen2025attribution,chen2025lifelong}.
As illustrated in Figure~\ref{fig:introduction}\textcolor{reda}{(A)}, each edit request provides a visual-textual context and a target correction, and the editor must absorb such requests sequentially into the MLLM.

Compared with text-only LLM editing, MLLM editing faces additional challenges induced by multimodal heterogeneity~\cite{cheng2023edit,chen2025attribution,liu2026principled}.
Recent studies have shown that directly transferring text-oriented editors to MLLMs leads to two distinct failure modes: unreliability caused by \textbf{Cross-modal Conflict} and instability caused by \textbf{Inter-edit Interference}~\cite{more}.
First, \emph{cross-modal conflict} arises because visual and textual representations may have different statistical scales and energy distributions, causing shared update statistics to be dominated by one modality~\cite{shi2025dualedit,more}. 
Second, \emph{inter-edit interference} emerges when a stream of edits repeatedly modifies overlapping parameter subspaces, resulting in long-horizon drift and forgetting~\cite{li2024can,more}. 
Modality-decoupled recursive editors mitigate these issues by separating modality-specific update statistics and performing stable Recursive writes in a fixed orthogonal low-rank edit subspace~\cite{more}.

Beyond reliability and stability, MLLM editing must also account for generality and locality~\cite{Huang2024vlkeb,fang2025can,du2025mmke}. 
As shown in Figure~\ref{fig:introduction}\textcolor{reda}{(B)}, existing editors mainly ask whether an edit succeeds on the original request and whether the model remains stable after many edits, but they do not explicitly define the semantic scope in which the edited knowledge should take effect. This omission leads to two scope-related errors. An edit may \emph{under-generalize}, failing to transfer to semantically equivalent questions or visual variants within the intended scope. It may also \emph{over-generalize}, leaking to unrelated inputs and changing behaviors that should remain intact.

To address this issue, we formalize \textbf{Edit-Scoped Generalization}, as illustrated in Figure~\ref{fig:introduction}\textcolor{reda}{(C)}. The core principle is that an edit should generalize only inside its intended semantic region and remain invisible outside it. This notion consists of two complementary aspects: \textbf{(1) In-scope Cross-modal Generalization}, where the edited knowledge transfers to semantically valid multimodal contexts, including other modalities that provide equivalent evidence~\cite{zhang-etal-2025-mc,NEURIPS2025_bbdd7f4a,jia-etal-2025-exploring}; and 
\textbf{(2) Out-of-scope Locality Preservation}, where the edited knowledge is prevented from leaking into inputs not semantically covered by the edit.

To probe whether reliable edits are truly scope-correct, we conduct pilot analyses of post-edit behaviors and internal MLLM activations (see Section~\ref{sec:edit_scoped_generalization}). 
Restricting the analysis to edits that already succeed on the original request, we find two complementary scope errors: $28.60\%$ of reliable edits fail to generalize to cross-modal variants within the edit scope, while $7.20\%$ leak into out-of-scope inputs and alter unrelated behaviors. 
Thus, reliability alone does not guarantee edit-scoped generalization~\cite{zhou-etal-2025-m2edit}.
We further observe that in-scope cross-modal generalization is not uniformly diffused throughout the network~\cite{jia-etal-2025-exploring}. 
Edit-related responses concentrate in deeper language layers, where target text and visual tokens show synchronized energy peaks; attention visualizations also reveal stronger post-edit attention from target text tokens to semantically relevant image regions~\cite{wang-etal-2025-v}. 

Motivated by these findings, we propose \textbf{ScopeEdit}, an online multimodal editor designed for controlling edit-scoped generalization. 
ScopeEdit separates each update into two functional components: a private absorption branch for reliable and stable modality-local writing, and a shared propagation branch for in-scope cross-modal generalization. 
The shared branch is activated only when visual and textual evidence are both directionally aligned and comparably supported, thereby enabling warranted cross-modal transfer while suppressing unwarranted leakage~\cite{Das_2026_CVPR,liu-etal-2025-insight,shao-etal-2025-cognition}.
To maintain long-horizon online efficiency and prevent inter-branch entanglement, both branches operate in orthogonal low‑rank subspaces derived from the partitioned rank budget, with each branch updated through its own recursive preconditioner. Our contributions are summarized as follows:
\begin{itemize}[leftmargin=*, labelsep=1.0em, noitemsep,nolistsep]
    % \item We introduce \textbf{Edit-Scoped Generalization} as a principled criterion for online multimodal knowledge editing. 
    % Our pilot analyses reveal that edit reliability does not imply scope correctness: reliable edits may either fail to transfer to semantically valid cross-modal variants or leak into unrelated inputs (Section~\ref{sec:prelim}).
    \item We introduce \textbf{Edit-Scoped Generalization} as a principled criterion for online multimodal knowledge editing that distinguishes warranted cross-modal transfer from unintended behavioral leakage. Pilot analyses show that reliability alone is insufficient, as reliable edits can still exhibit both in-scope under-generalization and out-of-scope over-generalization (Section~\ref{sec:prelim}).
    \item We propose \textbf{\lsy{ScopeEdit}}, a scope-aware online editor that decomposes each update into a modality-local absorption branch and an evidence-gated shared generalization branch. 
    By combining scope-separated low-rank write geometries, cross-modal evidence gating, and branch-wise recursive preconditioners, \lsy{ScopeEdit} enables warranted cross-modal transfer while suppressing unintended leakage with constant per-edit overhead (Section~\ref{sec:method}).
    \item We provide a comprehensive empirical evaluation across controlled benchmarks, long edit streams, real-world VLKEB scenarios, and diverse MLLM backbones. 
    The results show that \lsy{ScopeEdit} consistently improves edit-scoped behavior, achieving a favorable balance among reliable correction, in-scope cross-modal transfer, out-of-scope locality, long-horizon stability, and online efficiency (Section~\ref{sec:exps}).
\end{itemize}

\section{Related Works}
\label{sec:2}
This section reviews prior work related to our study. 
Sections~\ref{sec:related_llm_editing} and~\ref{sec:related_mllm_editing} summarize knowledge editing methods for LLMs and MLLMs, highlighting the shift from text-only editing toward multimodal editing. 
Sections~\ref{sec:related_online_llm} and~\ref{sec:related_online_mllm} further discuss online editing scenarios, where studies on MLLMs remain preliminary.

\subsection{Knowledge Editing for LLMs}
\label{sec:related_llm_editing}

Knowledge editing for text-only LLMs has been widely studied and can be broadly divided into \emph{parameter-modifying} and \emph{parameter-preserving} paradigms~\cite{dai-etal-2022-knowledge,DBLP:conf/iclr/SinitsinPPPB20,zhang2024comprehensive,DBLP:conf/iclr/HuangSZZR023}. 
Parameter-modifying methods directly alter model weights~\cite{pmlr-v235-ma24h,DBLP:conf/iclr/ZhangYLWRC25,ICLR2025_35cb54b8,zhai-etal-2025-parameter,yao-etal-2025-cake}, including \emph{locate-then-edit} approaches such as ROME~\cite{meng2022locating} and MEMIT~\cite{meng2023massediting}, as well as their extensions such as AlphaEdit~\cite{fangalphaedit} and DeltaEdit~\cite{cao2025deltaedit}. 
Another line of work uses \emph{meta-learning} to predict edits from gradients or error signals, such as KE~\cite{de-cao-etal-2021-editing} and MEND~\cite{DBLP:conf/iclr/MitchellLBFM22}. 
In contrast, parameter-preserving methods avoid changing the backbone~\cite{scialanga-etal-2025-sake,wang-etal-2025-knowledge-editing} and instead rely on retrieval, in-context editing, or lightweight auxiliary modules to override behavior locally, such as IKE~\cite{DBLP:conf/emnlp/ZhengLDFWXC23}, SERAC~\cite{mitchell2022memory}, and GRACE~\cite{DBLP:conf/nips/HartvigsenSPKG23}. 
These methods are effective in text-only regimes, but their assumptions about homogeneous representations and edit scope become fragile once visual inputs are introduced~\cite{chen2025attribution,DBLP:conf/aaai/YuCZH24}.

\subsection{Knowledge Editing for MLLMs}
\label{sec:related_mllm_editing}

Knowledge editing for MLLMs is still emerging~\cite{cheng2023edit}. 
Early studies mainly adapt LLM editors to multimodal benchmarks~\cite{he2024llmsmeetmultimodalgeneration,zhang2024comprehensive}, while MMEdit shows that such direct transfer is often brittle in coupled vision-language representations~\cite{cheng2023edit}.
Subsequent methods introduce multimodal-specific designs, such as attribution-guided visual editing in VisEdit~\cite{chen2025attribution} and gated dual-branch updating in DualEdit~\cite{shi2025dualedit}. 
More recent work, such as M-ORE~\cite{more} further highlights \emph{cross-modal conflict} as a central obstacle to multimodal editing, and emphasizes modality decoupling as a way to improve editing reliability. However, edit reliability only verifies the corrected behavior on the original request; it overlooks edit-scoped generalization~\cite{fang2025can,pmlr-v267-guo25e}.
Orthogonal to reliability-oriented improvements, recent \emph{parameter-preserving} methods have also started to explicitly model the scope of edit application.
For instance, MSCKE~\cite{zeng2025visual} extends SERAC~\cite{mitchell2022memory} with an external multimodal scope classifier to route only in-scope image-text queries to the edited memory.
This requirement becomes more stringent for \emph{parameter-modifying} editors, where the edit is absorbed into model weights and must simultaneously generalize to in-scope queries while avoiding out-of-scope leakage~\cite{zeng2025visual,more}.

\subsection{Online Editing for LLMs}
\label{sec:related_online_llm}

Online (or sequential) editing extends one-shot correction to a stream of updates~\cite{ICLR2025_2f89a23a}, requiring editors to continuously balance plasticity and stability over time~\cite{mitchell2022memory,jiang-etal-2025-neuron,pmlr-v235-wang24s,li-chu-2025-adaedit}. 
In text-only LLMs, retrieval- or memory-based methods alleviate catastrophic forgetting by storing past edits externally, but usually incur inference costs that grow with the number of edits~\cite{chen-etal-2024-lifelong,DBLP:conf/nips/0104L0XY0X0C24}. 
Parameter-modifying approaches are more efficient, yet often degrade under long edit streams due to accumulated interference and drift~\cite{meng2022locating,3737916.3738887}. 

\subsection{Online Editing for MLLMs}
\label{sec:related_online_mllm}

Online editing for MLLMs remains even less explored and is further complicated by multimodal heterogeneity~\cite{chen2025lifelong,chen2025attribution,shi2025dualedit}.
Existing methods are still dominated by parameter-preserving strategies. For example, LiveEdit~\cite{chen2025lifelong} formulates lifelong multimodal editing through edit-specific low-rank experts and inference-time routing, improving long-horizon robustness at the cost of edit-growing memory overhead~\cite{DBLP:conf/icml/ThedeRBAH25}.
In contrast, recent parameter-modifying work identifies \emph{inter-edit interference} as a core challenge and improves long-horizon stability via fixed orthogonal low-rank recursive writes under constant overhead~\cite{more}.

\section{Analysis on Multimodal  Knowledge Edit-Scoped Generalization}
\label{sec:analysis}

This section introduces the preliminary concepts and analysis necessary for our method. 
Section~\ref{sec:prelim} reviews the MLLM architecture, the FFN key-value view, and the online editing formulation with reliability, generality, and locality criteria. 
Section~\ref{sec:edit_scoped_generalization} analyzes \emph{edit-scoped generalization} in online MLLM editing, showing that reliable edits may still fail on in-scope variants or leak to out-of-scope inputs. 
Finally, we examine where cross-modal generalization emerges in MLLMs, providing empirical guidance for our method design.

\subsection{Preliminary}
\label{sec:prelim}
\paragraph{Multimodal Large Language Model}
We consider an edited MLLM $f_\theta$ composed of a vision encoder $\mathcal{E}^v$, a projector $\mathcal{P}$, and a language model $\mathcal{M}$. 
Given a visual input $x^v$ and a textual query $x^q$, the model autoregressively predicts the output distribution $p(y \mid x^v, x^q)$ from the fused token embeddings
$
H = [\mathcal{P}(\mathcal{E}^v(x^v)), \mathrm{Embed}(x^q)].
$
For a transformer layer $l$, we follow the standard residual decomposition~\cite{meng2022locating}:
\begin{equation}
h^{l}=h^{l-1}+a^{l}+m^{l}, \quad
m^{l}=W_{\mathrm{out}}^{l}\,\sigma\!\big(W_{\mathrm{in}}^{l}\,\gamma(h^{l-1}+a^{l})\big),
\end{equation}
where $a^{l}$ and $m^{l}$ denote the attention and feed-forward outputs, $\gamma(\cdot)$ is LayerNorm, and $\sigma(\cdot)$ is a nonlinearity. 
Following the FFN key-value interpretation of transformer memories~\cite{geva2021transformer,fangalphaedit}, we define
\begin{equation}
k^{l}\triangleq \sigma\!\big(W_{\mathrm{in}}^{l}\,\gamma(h^{l-1}+a^{l})\big),\quad
v^{l}\triangleq m^{l}=W_{\mathrm{out}}^{l}k^{l}.
\end{equation}
Here $k^{l}$ acts as a \emph{key} computed from the current multimodal context, and $v^{l}$ is the \emph{value} retrieved linearly by $W_{\mathrm{out}}^{l}$. Under this view, many \emph{parameter-modifying} editing methods intervene on $W_{\mathrm{out}}^{l}$ to rewrite selected $k \rightarrow v$ associations. For simplicity, we use $W$ to refer to $W_{\mathrm{out}}^{l}$ in the following.

\begin{figure*}
    \centering
    \includegraphics[width=1.0 \linewidth]{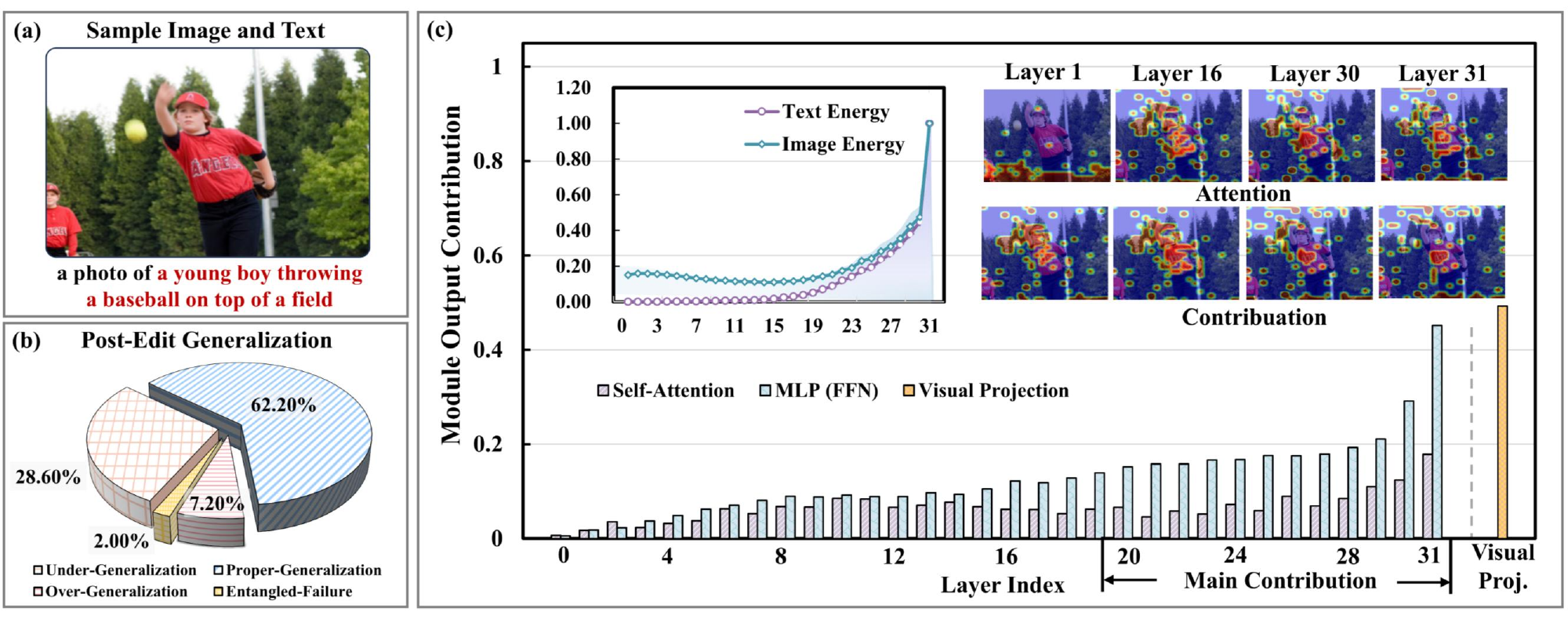}
\caption{Pilot analysis of edit-scoped generalization and its layer-wise emergence in online MLLM editing, using M-ORE on LLaVA-v1.5. 
\textbf{(a)} An example edit sample with paired image and text. 
\textbf{(b)} Reliable edits may still exhibit under-generalization, over-generalization, or entangled failure. 
\textbf{(c)} Edit-related cross-modal responses mainly emerge in deeper language layers, where textual and visual evidence becomes more synchronized. 
More discussions on other MLLM architectures are provided in Section~\ref{sec:cross_arch_extension}.}
    \label{fig:pilot-study}
    \vspace{-3mm}
\end{figure*}

\paragraph{Online Editing in MLLMs}
\label{sec:online}
We consider online editing of an MLLM $f_\theta:\mathcal{X}\rightarrow\mathcal{Y}$ that maps a multimodal input $x^e=(x^v,x^q)\in\mathcal{X}$ to a textual output distribution $y^e$. Following the key-value view above, we focus on \emph{parameter-modifying} edits that intervene on the selected FFN output matrix $W$, while keeping the remaining parameters frozen. Concretely, we partition parameters as $\theta=(\theta_0,W)$, where $\theta_0$ is fixed and only $W$ is editable. An edit request at step $t$ is specified by a target input-output pair $e_t=(x_t^e,y_t^e)$, where $y_t^e\neq f_{(\theta_0,W_{t-1})}(x_t^e)$. Starting from a base model with editable matrix $W_0$, a model editor $\mathrm{ME}$ performs \emph{online sequential} edits by producing an additive update $\Delta W_t$ with a bounded per-edit cost:
\begin{equation}
\Delta W_t=\mathrm{ME}(f_{(\theta_0,W_{t-1})},x_t^e,y_t^e),\quad
W_t=W_{t-1}+\Delta W_t.
\end{equation}
Equivalently, the edited model at step $t$ is $f_{\theta_t}=f_{(\theta_0,W_t)}$ with
$
W_t=W_0+\sum_{i=1}^{t}\Delta W_i.
$
A successful editor should satisfy three criteria at each step $t$, as outlined in~\cite{cheng2023edit,chen2025lifelong}:

\textbf{Reliability} requires the model to produce the desired answer on the edited input:
\begin{equation}
\mathbb{E}_{(x_t^e,y_t^e)}
\big[
\mathbb{I}(f_{\theta_t}(x_t^e)=y_t^e)
\big],
\end{equation}
where $\mathbb{I}$ is a correctness indicator used to compute accuracy.

\textbf{Generality} 
measures whether the edit holds under semantically equivalent variants of the request, including both textual and visual perturbations that fall within the edit scope. 
This captures \emph{in-scope cross-modal generalization}, where the edited knowledge is expected to transfer to contexts providing equivalent semantic evidence. 
Let $\mathcal{N}_t(x_t^e)$ be such an in-scope neighborhood, e.g., paraphrases of $x_t^q$ or benign transformations of $x_t^v$; then:
\begin{equation}
\mathbb{E}_{x'\sim\mathcal{N}_t(x^e_t)}
\big[
\mathbb{I}(f_{\theta_t}(x')=y_e^t)
\big].
\end{equation}

\textbf{Locality} 
demands minimal side effects on unrelated inputs outside the edit scope. 
This captures \emph{out-of-scope locality preservation}, requiring the edited knowledge to remain invisible to samples not semantically connected to the edit. 
Let $\mathcal{U}_t$ denote such an irrelevant set; we measure distribution shift by the KL divergence between the next-token distributions of the post-edit and pre-edit models:
\begin{equation}
\mathbb{E}_{x\sim\mathcal{U}_t}
\left[
\exp\!\left(
-\mathrm{KL}\!\big(
p_{\theta_t}(\cdot|x)\,\|\,p_{\theta_{t-1}}(\cdot|x)
\big)
\right)
\right],
\end{equation}
where $p_\theta(\cdot|x)$ is the next-token distribution under the same decoding prefix. 
% Detailed full  definitions are provided in \textbf{Appendix~\ref{app:rgl}}. 
Next, we conduct a pilot study on online MLLM editing to analyze \emph{edit-scoped generalization} and derive the design principles of our method.

\subsection{Edit-Scoped Generalization}
\label{sec:edit_scoped_generalization}

\subsubsection{Successful Edits Do Not Imply Scope-Correct Generalization}

We restrict our analysis to edits that are reliable on the original edited request, excluding failed edit cases~\footnote{Following the online editing protocol of M-ORE, we record reliability $\mathrm{Rel}_t$, text/image generality $T\text{-}\mathrm{Gen}_t$ and $M\text{-}\mathrm{Gen}_t$, and text/image locality $T\text{-}\mathrm{Loc}_t$ and $M\text{-}\mathrm{Loc}_t$ after each edit. We keep edits with $\mathrm{Rel}_t\ge\tau_{\mathrm{rel}}$, and set $\mathrm{Gen}_t=\min(T\text{-}\mathrm{Gen}_t,M\text{-}\mathrm{Gen}_t)$ and $\mathrm{Loc}_t=\min(T\text{-}\mathrm{Loc}_t,M\text{-}\mathrm{Loc}_t)$. Given thresholds $\tau_{\mathrm{gen}}$ and $\tau_{\mathrm{loc}}$, reliable edits are categorized as  \textsc{Proper Propagation} if $\mathrm{Gen}_t\ge\tau_{\mathrm{gen}},\mathrm{Loc}_t\ge\tau_{\mathrm{loc}}$, \textsc{Under-Propagation} if $\mathrm{Gen}_t<\tau_{\mathrm{gen}},\mathrm{Loc}_t\ge\tau_{\mathrm{loc}}$, \textsc{Over-Propagation} if $\mathrm{Gen}_t\ge\tau_{\mathrm{gen}},\mathrm{Loc}_t<\tau_{\mathrm{loc}}$, and \textsc{Entangled Failure} if $\mathrm{Gen}_t<\tau_{\mathrm{gen}},\mathrm{Loc}_t<\tau_{\mathrm{loc}}$. We set $\tau_{\mathrm{rel}}=1.0$, $\tau_{\mathrm{gen}}=0.8$, and $\tau_{\mathrm{loc}}=0.95$.}. 
As shown in Figure~\ref{fig:pilot-study}\textcolor{reda}{(b)},
among reliable edits, only $62.20\%$ achieve proper generalization, while the remaining cases suffer from different scope errors. 
Specifically, $28.60\%$ exhibit under-generalization, failing to transfer to in-scope cross-modal variants; $7.20\%$ exhibit over-generalization, leaking to out-of-scope inputs; and $2.00\%$ involve entangled failures. 
This shows that reliability alone cannot determine the semantic boundary of an edit. 
A practical online MLLM editor should therefore not only write the corrected knowledge, but also control its propagation scope: generalizing to semantically warranted multimodal variants while remaining inactive on unrelated inputs.
% These results motivate that generalization in MLLM editing should be edit-scoped: an edit should generalize where it is semantically warranted, while remaining invisible elsewhere.

\subsubsection{In-Scope Generalization Emerges in Late Shared Semantic Layers}
We further analyze where edit-scoped cross-modal generalization emerges within MLLMs. 
Specifically, we quantify layer-wise edit responses for target text tokens and visual tokens, and compare their energy profiles across MLLM layers. 
As shown in Figure~\ref{fig:pilot-study}\textcolor{reda}{(c)}, edit-related responses are not uniformly distributed throughout the network. 
Instead, the strongest responses concentrate in deeper language layers, where the response peaks of target text and visual tokens become increasingly synchronized. 
We also provide qualitative attention visualizations in Figure~\ref{fig:pilot-study}\textcolor{reda}{(c)}. After editing, target text tokens attend more strongly to semantically relevant image regions in deeper layers.

\section{Methodology}
\label{sec:method}

Motivated by the above observations, we formulate online MLLM editing as a scope-aware two-branch update framework: stable modality-local absorption and evidence-gated shared generalization. The former enables reliable editing while \emph{preserving out-of-scope locality}, whereas the latter activates \emph{in-scope cross-modal generalization} only when supported by sufficient semantic evidence in deeper layers, thereby preventing the edit from propagating beyond its intended scope.

\subsection{Edit-Scoped Online Editing Principle}
\label{sec:edit_scoped_principle}
Existing online editors for MLLMs can be viewed as proximal updates under history-dependent geometric regularization~\cite{more}. 
At each step, given an edit sample $(x_t^v,x_t^q,y_t)$, the editor incorporates the current correction by minimizing
$\mathcal{L}_{\mathrm{edit}}^t
=
\mathbb{E}_{(x_t^v,x_t^q,y_t)}
[-\log p_\theta(y_t \mid x_t^v,x_t^q)]$,
while constraining the update with accumulated second-order statistics, so as to reduce interference with previously edited and unrelated knowledge. 
Formally, let $W_{t-1}$ denote the editable parameter at step $t$, and let
$
g_t=\nabla_{W_{t-1}}\mathcal{L}_{\mathrm{edit}}^t
$
be the edit gradient.
The update can be written as
\begin{equation}
\label{eq:proximal}
\Delta W_t
=
\arg\min_{\Delta W}
\left\|
\Delta W + \eta g_t
\right\|_F^2
+
\operatorname{Tr}\!\left(
\Delta W C_{t-1} \Delta W^\top
\right),
\end{equation}
where $C_{t-1}$ denotes the second-order statistic accumulated from the edit history. 
Specifically, at each previous step $s<t$, we collect a fixed-size \emph{step-only} context set $\mathcal{B}_s$, consisting of the edit batch and a small locality sample at that step. 
Then,
\begin{equation}
\small
C_{t-1}
=
C_0
+
\sum_{s=1}^{t-1}
\sum_{x\in\mathcal{B}_s}
k_s(x)k_s(x)^\top,
\quad
C_0=0,
\label{eq:full_space_C}
\end{equation}
where $k_s(x)$ is the FFN key induced by multimodal input $x$ at the edited module. 
By construction, $C_{t-1}$ excludes any context from the current edit step $t$ and therefore serves as a purely historical geometry for the current write.

When $C_{t-1} \succeq 0$, Eq.~\eqref{eq:proximal} is strictly convex and admits the closed-form solution:
\begin{equation}
\Delta W_t
=
-\eta\, g_t \big(I+C_{t-1}\big)^{-1}
\triangleq
-\eta\, g_t P_{t-1},
\label{eq:prox_closed_form}
\end{equation}
where $P_{t-1} \triangleq (I+C_{t-1})^{-1}$ serves as a history-dependent preconditioner. 
This view interprets online editing as a gradient write shaped by a stability-history geometry.

However, the above objective does not explicitly model the generalizability of edits. As shown in Section~\ref{sec:edit_scoped_generalization}, a reliable edit may fail to transfer to in-scope cross-modal variants, or may leak to out-of-scope inputs and thereby compromise locality. To control this behavior, we decompose the update at each layer into two functionally separated branches:
\begin{equation}
\small
\Delta W_t^{(l)}
=
\Delta W_{t,\mathrm{loc}}^{(l)}
+
\Delta W_{t,\mathrm{sh}}^{(l)},
\label{eq:two_branch_decomp}
\end{equation}
where $\Delta W_{t,\mathrm{loc}}^{(l)}$ denotes the modality-local absorption branch, and $\Delta W_{t,\mathrm{sh}}^{(l)}$ denotes the shared generalization branch. 
The local branch is always active and is responsible for reliable editing while preserving out-of-scope locality. 
The shared branch is activated only when the current edit provides sufficient cross-modal semantic evidence, enabling in-scope cross-modal generalization.

Accordingly,  we instantiate Eq.~\eqref{eq:proximal} and derive the following edit-scoped proximal objective:
\begin{equation}
\label{eq:edit_scoped_objective}
\resizebox{1.0\columnwidth}{!}{$
\begin{aligned}
\min_{\Delta W_{t, \mathrm{loc}}^{(l)},\,\Delta W_{t, \mathrm{sh}}^{(l)}}
\quad
&
\left\|
\Delta W_{\mathrm{loc}}^{(l)}
+
\eta_{\mathrm{loc}} g_{t,\mathrm{loc}}^{(l)}
\right\|_F^2
+
\operatorname{Tr}\!\left(
\Delta W_{\mathrm{loc}}^{(l)}
C_{{t-1},\mathrm{loc}}^{(l)}
{\Delta W_{\mathrm{loc}}^{(l)}}^\top
\right)
\\
&+
\left\|
\Delta W_{\mathrm{sh}}^{(l)}
+
\eta_{\mathrm{sh}} \gamma_t^{(l)} g_{t,\mathrm{sh}}^{(l)}
\right\|_F^2
+
\operatorname{Tr}\!\left(
\Delta W_{\mathrm{sh}}^{(l)}
C_{{t-1},\mathrm{sh}}^{(l)}
{\Delta W_{\mathrm{sh}}^{(l)}}^\top
\right).
\end{aligned}
$}
\end{equation}

Furthermore, $C_{t-1}$ in Eq.~\eqref{eq:full_space_C} is instantiated as branch-specific edit-scope-aware historical statistics.
The local branch accumulates FFN keys from the textual context, whereas the shared branch accumulates gated cross-modal keys constructed from the visual input and textual query:
\begin{equation}
\small
C_{t-1,\mathrm{loc}}^{(l)}
=
C_{0,\mathrm{loc}}^{(l)}
+
\sum_{s=1}^{t-1}
\sum_{(x^v,x^q)\in\mathcal{B}_s}
k_{s}^{(l)}(x^q)
{k_{s}^{(l)}(x^q)}^\top,
\label{eq:C_loc_full}
\end{equation}
\begin{equation}
\small
C_{t-1,\mathrm{sh}}^{(l)}
=
C_{0,\mathrm{sh}}^{(l)}
+
\sum_{s=1}^{t-1}
\gamma_s^{(l)}
\sum_{(x^v,x^q)\in\mathcal{B}_s}
k_{s}^{(l)}(x^v,x^q)
{k_{s}^{(l)}(x^v,x^q)}^\top,
\label{eq:C_sh_full}
\end{equation}
where $\gamma_t^{(l)}\in[0,1]$ is a scope gate for the shared branch. 
It scales the current shared write in Eq.~\eqref{eq:edit_scoped_objective} and, analogously, $\gamma_s^{(l)}$ weights the contribution of historical step $s$ in Eq.~\eqref{eq:C_sh_full}. 
Thus, only edits supported by sufficient cross-modal semantic evidence are allowed to enter and shape the shared geometry, enabling in-scope generalization while suppressing out-of-scope propagation.

\subsection{Scope-Separated Write Geometry}
\label{sec:scope_write_geometry}

To prevent entanglement between local absorption and shared generalization within the same parameter space while satisfying online editing efficiency, we employ a fixed low-rank write interface and partition the rank budget of each editable layer into two orthogonal branches.
Let the total rank be $r=r_\ell+r_s$. We parameterize the layer-wise update as
\begin{equation}
\small
\Delta W_t^{(l)}
=
\Delta W_{t,\mathrm{loc}}^{(l)}
+
\Delta W_{t,\mathrm{sh}}^{(l)} = 
\Delta B_{t,\mathrm{loc}}^{(l)} A_{\mathrm{loc}}^{(l)}
+
\Delta B_{t,\mathrm{sh}}^{(l)} A_{\mathrm{sh}}^{(l)},
\label{eq:scope_write_decomp}
\end{equation}
where 
$\Delta B_{t,\mathrm{loc}}^{(l)}$
and
$\Delta B_{t,\mathrm{sh}}^{(l)}$
are zero-initialized and updated recursively along the edit stream.
$
A_{\mathrm{loc}}^{(l)} \in \mathbb{R}^{r_\ell \times d}
$
and
$
A_{\mathrm{sh}}^{(l)} \in \mathbb{R}^{r_s \times d}
$
denote the fixed low-rank write coordinates.
To prevent the two effects from being geometrically entangled, we impose an orthogonal coordinate split:
\begin{equation}
A_{\mathrm{loc}}^{(l)} {A_{\mathrm{loc}}^{(l)}}^\top = I,
\quad
A_{\mathrm{sh}}^{(l)} {A_{\mathrm{sh}}^{(l)}}^\top = I,
\quad
A_{\mathrm{loc}}^{(l)} {A_{\mathrm{sh}}^{(l)}}^\top = 0.
\label{eq:shared_private_orth}
\end{equation}

Notably, after orthogonal initialization, $A_{\mathrm{loc}}^{(l)}$ and $A_{\mathrm{sh}}^{(l)}$ remain fixed throughout online editing. 

Correspondingly, each FFN key is projected into its branch-specific write subspace through a unified projection rule:
\begin{equation}
\small
z_{b}^{(l)}(u)
=
A_{b}^{(l)} k^{(l)}(u),
\quad
b\in\{\mathrm{loc},\mathrm{sh}\},
\label{eq:generic_key_projection}
\end{equation}
where $u$ denotes the branch-dependent input context used to induce the FFN key. 
In practice, the local branch instantiates $u$ as $x^q$, while the shared branch instantiates $u$ as $(x^v,x^q)$. 
Applying this projection to the historical keys in Eqs.~\eqref{eq:C_loc_full} and~\eqref{eq:C_sh_full} directly yields the corresponding low-rank branch-wise geometries used for online preconditioning.

\subsection{Scope Activation from Cross-Modal Semantic Evidence}
\label{sec:scope_activation}

Cross-modal generalization is desirable only within the edit scope; beyond this scope, it becomes leakage that harms \emph{out-of-scope locality preservation}. 
Therefore, we activate the shared generalization branch only when the current visual and textual evidence jointly supports the same edited semantics. 
When such evidence is insufficient or inconsistent, the current edit is mainly absorbed by the modality-local branch.

At edit step $t$, for each candidate shared layer $l$,  we extract two FFN keys from the current edit input $ (x^v,x^q)$.
Following the projection rule in Eq.~\eqref{eq:generic_key_projection}, both keys are mapped into the shared write subspace:
\begin{equation}
z_{t,\mathrm{text}}^{(l)} = A_{\mathrm{sh}}^{(l)} k_t^{(l)}(x^q),
\quad
z_{t,\mathrm{vision}}^{(l)} = A_{\mathrm{sh}}^{(l)} k_t^{(l)}(x^v).
\label{eq:scope_project}
\end{equation}

We characterize the cross-modal evidence with two complementary factors: directional agreement and bilateral support~\cite{chen2025lifelong}. 
Directional agreement measures whether the two modalities point to a consistent semantic direction in the shared space, while bilateral support measures whether the evidence is comparably supported by both modalities:
\begin{align}
\small
\operatorname{cos}^{(l)}_t
&=
\frac{
\left\langle z_{t,\mathrm{text}}^{(l)},\,z_{t,\mathrm{vision}}^{(l)} \right\rangle
}{
\|z_{t,\mathrm{text}}^{(l)}\|_2
\,
\|z_{t,\mathrm{vision}}^{(l)}\|_2
+
\varepsilon
},
\\
\operatorname{sup}^{(l)}_t
&=
\frac{
\min\!\big(
\|z_{t,\mathrm{text}}^{(l)}\|_2,
\|z_{t,\mathrm{vision}}^{(l)}\|_2
\big)
}{
\max\!\big(
\|z_{t,\mathrm{text}}^{(l)}\|_2,
\|z_{t,\mathrm{vision}}^{(l)}\|_2
\big)
+
\varepsilon
}.
\end{align}

Based on these two quantities, we define the scope gate as
\begin{equation}
\small
\gamma_t^{(l)}
=
\sigma\!\left(
\beta
\left(
\operatorname{cos}^{(l)}_t-\tau
\right)
\right)
\cdot
\operatorname{sup}^{(l)}_t,
\label{eq:gamma_gate}
\end{equation}
where $\tau$ is the activation threshold and $\beta$ controls the gate sharpness. 
The gate becomes large only when the textual and visual evidence are directionally aligned and comparably supported. 
Thus, as shown in Eq.~\eqref{eq:edit_scoped_objective} and Eq.~\eqref{eq:C_sh_full}, $\gamma_t^{(l)}$ suppresses shared propagation under cross-modal conflicts or weak visual-textual evidence, while activating it when the edit is likely to generalize within the intended multimodal scope.

\begin{algorithm}[t]
\caption{Online Editing Pipeline of \lsy{ScopeEdit}}
\label{alg:eva}
\small

\setlength{\abovedisplayskip}{2pt}
\setlength{\belowdisplayskip}{2pt}
\setlength{\abovedisplayshortskip}{1pt}
\setlength{\belowdisplayshortskip}{1pt}
\setlength{\jot}{1pt}

\begin{algorithmic}[1]
\setlength{\itemsep}{0.15em}
\setlength{\parsep}{0pt}
\setlength{\partopsep}{0pt}
\setlength{\topsep}{0pt}
\REQUIRE Base MLLM $f_{\theta_0}$, edit stream $\{e_t=(x_t^e,y_t^e)\}_{t=1}^{T}$ with $x_t^e=(x_t^v,x_t^q)$, edited layers $\mathcal{L}$, branch ranks $r_\ell,r_s$, and  $\eta$
\ENSURE Edited model sequence $\{f_{\theta_t}\}_{t=1}^{T}$

\STATE Construct fixed orthogonal bases $\{A_{\mathrm{loc}}^{(l)},A_{\mathrm{sh}}^{(l)}\}_{l\in\mathcal{L}}$
\FOR{$l\in\mathcal{L}$}
    \STATE Initialize branch coefficients $B_{\mathrm{loc}}^{(l)}\leftarrow 0$, $B_{\mathrm{sh}}^{(l)}\leftarrow 0$
    \STATE Initialize recursive preconditioners $P_{\mathrm{loc}}^{(l)}\leftarrow I_{r_\ell}$, $P_{\mathrm{sh}}^{(l)}\leftarrow I_{r_s}$
\ENDFOR

\FOR{$t=1,\ldots,T$}
    \STATE Receive current edit request $e_t=(x_t^v, x_t^q, y_t^e)$
    \STATE Compute edit loss $\mathcal{L}_{\mathrm{edit}}(f_{\theta_{t-1}}(x_t^v, x_t^q),y_t^e)$

    \FOR{$l\in\mathcal{L}$}
        \STATE Extract branch-specific FFN keys $k_{t,\mathrm{loc}}^{(l)}$ and $k_{t,\mathrm{sh}}^{(l)}$
        \STATE Project keys into scope-separated subspaces:
        \[
        z_{t,\mathrm{loc}}^{(l)}=A_{\mathrm{loc}}^{(l)}k_{t,\mathrm{loc}}^{(l)},\quad
        z_{t,\mathrm{sh}}^{(l)}=A_{\mathrm{sh}}^{(l)}k_{t,\mathrm{sh}}^{(l)}.
        \]
        \STATE Estimate cross-modal evidence gate $\gamma_t^{(l)}$
        \STATE Set branch weights $\rho_{t,\mathrm{loc}}^{(l)}=1$ and $\rho_{t,\mathrm{sh}}^{(l)}=\gamma_t^{(l)}$

        \FOR{$b\in\{\mathrm{loc},\mathrm{sh}\}$}
            \STATE Compute low-rank branch gradient:
            \[
            G_{t,b}^{(l)}=\nabla_{B_b^{(l)}}\mathcal{L}_{\mathrm{edit}}.
            \]
            \STATE Apply history-preconditioned branch write:
            \[
            \Delta B_{t,b}^{(l)}
            =
            -\eta_b\,\rho_{t,b}^{(l)}
            G_{t,b}^{(l)}
            P_{t-1,b}^{(l)}.
            \]
            \STATE Update branch coefficient:
            \[
            B_b^{(l)}\leftarrow B_b^{(l)}+\Delta B_{t,b}^{(l)}.
            \]
        \ENDFOR
        \FOR{$b\in\{\mathrm{loc},\mathrm{sh}\}$}
            \STATE Form weighted current key:
            \[
            \widetilde z_{t,b}^{(l)}
            =
            \sqrt{\rho_{t,b}^{(l)}}\,z_{t,b}^{(l)}.
            \]
            \STATE Update recursive preconditioner by Sherman--Morrison:
            \[
            P_{t,b}^{(l)}
            =
            P_{t-1,b}^{(l)}
            -
            \frac{
            P_{t-1,b}^{(l)}
            \widetilde z_{t,b}^{(l)}
            (\widetilde z_{t,b}^{(l)})^\top
            P_{t-1,b}^{(l)}
            }{
            1+
            (\widetilde z_{t,b}^{(l)})^\top
            P_{t-1,b}^{(l)}
            \widetilde z_{t,b}^{(l)}
            }.
            \]
        \ENDFOR
    \ENDFOR

    \STATE Obtain edited model $f_{\theta_t}$ with updated matrices $\{W_t^{(l)}\}_{l\in\mathcal{L}}$
\ENDFOR
\end{algorithmic}
\end{algorithm}

\subsection{Branch-wise Recursive Online Updates}
\label{sec:dynamic_scope_routing}

We now derive the recursive online update rule in the scope-separated low-rank space. The edit at step $t$ proceeds in two stages: history-preconditioned branch writing, followed by post-write geometric recursion for future edits.

\paragraph{History-preconditioned branch write}
For each branch and layer, we maintain a compact inverse geometry:
\begin{equation}
\small
P_{t-1,b}^{(l)}
=
\left(
I+C_{t-1,b}^{(l)}
\right)^{-1},
\quad
b\in\{\mathrm{loc},\mathrm{sh}\}.
\label{eq:P_branch}
\end{equation}
Notably, the step-$t$ editing uses $P_{t-1,b}^{(l)}$, rather than $P_{t,b}^{(l)}$, ensuring that the current edit is constrained only by past edits.

Following Eqs.~\eqref{eq:prox_closed_form} and~\eqref{eq:scope_write_decomp}, we only need to recursively update the branch-wise write coefficients 
$\Delta B_{t,\mathrm{loc}}^{(l)}$
and
$\Delta B_{t,\mathrm{sh}}^{(l)}$. 
Let
$
G_{t,\mathrm{loc}}^{(l)}
$
and
$
G_{t,\mathrm{sh}}^{(l)}
$
denote the corresponding edit gradients.  
The resulting branch-wise closed-form updates are
\begin{equation}
\small
\Delta B_{t,\mathrm{loc}}^{(l)}
=
-\eta_{\mathrm{loc}}
G_{t,\mathrm{loc}}^{(l)}
P_{t-1,\mathrm{loc}}^{(l)},
\;
\Delta B_{t,\mathrm{sh}}^{(l)}
=
-\eta_{\mathrm{sh}}
\gamma_t^{(l)}
G_{t,\mathrm{sh}}^{(l)}
P_{t-1,\mathrm{sh}}^{(l)}.
\label{eq:B_branch}
\end{equation}

\paragraph{Post-write geometry recursion}
After the write, we compute the projected keys of the current edit step and incorporate them into the inverse geometries for future updates. 
For the current edit input $(x_t^v,x_t^q)$, we instantiate
$
z_{t,\mathrm{loc}}^{(l)}
=
z_{\mathrm{loc}}^{(l)}(x_{t}^q)
$
and
$
z_{t,\mathrm{sh}}^{(l)}
=
z_{\mathrm{sh}}^{(l)}(x_{t}^v,x_{t}^q)
$ using Eq.~\eqref{eq:generic_key_projection}.
To unify the local and shared recursions, we define
\begin{equation}
\small
\widetilde z_{t,b}^{(l)}
=
\sqrt{\rho_{t,b}^{(l)}}\,z_{t,b}^{(l)},
\quad
\rho_{t,\mathrm{loc}}^{(l)}=1,
\quad
\rho_{t,\mathrm{sh}}^{(l)}=\gamma_t^{(l)}.
\label{eq:weighted_branch_key}
\end{equation}

Using the Sherman--Morrison lemma, we recursively update the branch-specific inverse geometry for the next edit step $t+1$, thereby avoiding the cost of explicit matrix inversion:
\begin{equation}
\small
P_{t,b}^{(l)}
=
P_{t-1,b}^{(l)}
-
\frac{
P_{t-1,b}^{(l)}
\widetilde z_{t,b}^{(l)}
\left(\widetilde z_{t,b}^{(l)}\right)^\top
P_{t-1,b}^{(l)}
}{
1+
\left(\widetilde z_{t,b}^{(l)}\right)^\top
P_{t-1,b}^{(l)}
\widetilde z_{t,b}^{(l)}
},
\quad
b\in\{\mathrm{loc},\mathrm{sh}\},
\label{eq:SM_branch}
\end{equation}
where $P_{t-1,b}^{(l)}$ preconditions the current update $\Delta B_{t,b}^{(l)}$, while $P_{t,b}^{(l)}$ is cached and used only from step $t+1$ onward. 
The local geometry is updated by every edit, whereas the shared geometry is scaled by $\gamma_t^{(l)}$, allowing only edits with sufficient cross-modal evidence to substantially shape the shared preconditioner. Each active layer stores only $P_{t,b}^{(l)}\in\mathbb{R}^{r_b\times r_b}$ and updates it in $O(r_b^2)$ time, giving constant per-edit overhead with respect to the edit stream length. 

Algorithm~\ref{alg:eva} provides a detailed presentation of the overall editing pipeline, elucidating the workflow of our \lsy{ScopeEdit}.

\begin{proof}[\textbf{Derivation of Eq.~\eqref{eq:SM_branch}}]
For each editable layer $l$ and branch $b\in\{\mathrm{loc},\mathrm{sh}\}$, ScopeEdit maintains
$
P_{t,b}^{(l)}=(I+C_{t,b}^{(l)})^{-1}.
$
After the current editing, the projected key is inserted into the historical statistic:
\begin{equation}
\small
C_{t,b}^{(l)}
=
C_{t-1,b}^{(l)}
+
\widetilde z_{t,b}^{(l)}
\big(\widetilde z_{t,b}^{(l)}\big)^\top .
\end{equation}
Let
$
H_{t-1,b}^{(l)}=I+C_{t-1,b}^{(l)}
$
and
$
u=\widetilde z_{t,b}^{(l)}
$.
Then the next-step inverse geometry is
\begin{equation}
\small
P_{t,b}^{(l)}
=
\left(H_{t-1,b}^{(l)}+uu^\top\right)^{-1}.
\end{equation}
Applying the Sherman--Morrison identity,
\begin{equation}
\small
(H+uu^\top)^{-1}
=
H^{-1}
-
\frac{H^{-1}uu^\top H^{-1}}{1+u^\top H^{-1}u},
\end{equation}
with $H=H_{t-1,b}^{(l)}$ and $H^{-1}=P_{t-1,b}^{(l)}$ directly yields Eq.~\eqref{eq:SM_branch}. 
Therefore, the inverse geometry can be updated through a rank-one correction in the branch-specific low-rank space, avoiding explicit matrix inversion while keeping the online update independent of the edit stream length.
\end{proof}

\section{Experiments}
\label{sec:exps}

To evaluate the effectiveness, stability, efficiency, and extensibility of \lsy{ScopeEdit}, we organize experiments around five research questions, followed by a qualitative case study for intuitive analysis.
We first compare \lsy{ScopeEdit} with mainstream online editors to assess its ability to balance in-scope cross-modal generalization and out-of-scope locality preservation (Section~\ref{sec:rq1}). 
We then analyze its core components, basis initialization, and hyperparameter sensitivity (Section~\ref{sec:design_analysis}), followed by long-horizon stability analysis on inter-edit interference and representation drift (Section~\ref{sec:rq3}). 
We further examine its theoretical and empirical efficiency under online constraints (Section~\ref{sec:rq4}), and finally evaluate whether the pilot observations generalize across MLLM architectures and real-world or more complex vision-language scenarios (Section~\ref{sec:cross_arch_extension}). 
The research questions are summarized as follows:
\begin{itemize}[leftmargin=*, labelsep=0.8em, noitemsep,nolistsep]
    \item \textbf{RQ1:}
    How does \lsy{ScopeEdit} compare with existing mainstream editors in online MLLM editing, particularly in balancing \emph{in-scope cross-modal generalization} and \emph{out-of-scope locality preservation}?
    
    \item \textbf{RQ2:} How do the core components and configurations of \lsy{ScopeEdit} affect its edit-scoped performance, including branch-wise ablations, write-space basis initialization, and hyperparameter sensitivity?

    \item \textbf{RQ3:}
    Can \lsy{ScopeEdit} maintain stable edit-scoped behavior over long online edit streams, such as mitigating inter-edit interference and hidden representation drift?

    \item \textbf{RQ4:}
    What are the time and space complexities of \lsy{ScopeEdit} for each online edit, and how does it achieve a quality--efficiency trade-off under strict online constraints?

    \item \textbf{RQ5:}
    Are the phenomena revealed by our pilot study consistent across different MLLM architectures? Furthermore, can \lsy{ScopeEdit} be extended to real-world scenarios and more complex vision-language architectures?
\end{itemize}

\begin{table}[t]
  \centering
  \setlength{\tabcolsep}{3pt}
  \renewcommand{\arraystretch}{1.08}
  \caption{Statistics of datasets used for online editing evaluation.}
  \label{tab:dataset_stats}
  \resizebox{\linewidth}{!}{
  \begin{tabular}{l|ccc|ccc}
    \toprule
    \textbf{Dataset} 
    & \textbf{Train} 
    & \textbf{Test} 
    & \textbf{Images}
    & \textbf{Reliability} 
    & \textbf{T/M-Generality} 
    & \textbf{T/M-Locality} \\
    \midrule
    E-VQA & 6,346 & 2,093 & 1,390 & 2,093 & 2,093 / 2,093 & 4,289 / 5,046 \\
    E-IC  & 2,849 & 1,000 & 1,000 & 1,000 & 1,000 / 1,000 & 4,289 / 5,046 \\
    \bottomrule
  \end{tabular}
  }
  \vspace{-3mm}
\end{table}

\subsection{Experiment Setup}

\subsubsection{Datasets}
Following~\cite{cheng2023edit}, we evaluate online MLLM editing on E-VQA (Editing VQA) and E-IC (Editing Image Caption), where E-IC requires finer-grained visual grounding. 
Detailed dataset statistics are summarized in Table~\ref{tab:dataset_stats}.

\subsubsection{Metrics}
For E-VQA and E-IC, we report \emph{Reliability}, \emph{Generality}, and \emph{Locality} (Section~\ref{sec:prelim}), and further decompose Generality/Locality into text- and image-conditioned evaluations. 

\subsubsection{MLLM Backbones \& Baseline Editors}
For comprehensive evaluation, we select diverse MLLM backbones spanning architectures and parameter scales, including BLIP2-OPT (2.7B)~\cite{li2023blip} and LLaVA-v1.5 (7B)~\cite{DBLP:conf/cvpr/LiuLLL24}.
Given the lack of effective online multimodal editors, we follow~\cite{cheng2023edit} and align with the setup in Section~\ref{sec:prelim}. Specifically, we adapt LLM-based editors to MLLMs and categorize them into two groups: \textit{parameter-modifying} and \textit{parameter-preserving}. The former includes FT-L~\cite{cheng2023edit}, FT-M~\cite{cheng2023edit}, MEND~\cite{DBLP:conf/iclr/MitchellLBFM22}, AlphaEdit~\cite{fangalphaedit} and \lsy{M-ORE~\cite{more}}, while the latter covers IKE~\cite{DBLP:conf/emnlp/ZhengLDFWXC23}, SERAC~\cite{mitchell2022memory}, and LiveEdit~\cite{chen2025lifelong}.

\subsubsection{Implementation Details}
For FT-L, FT-M, MEND, SERAC, and LiveEdit, we follow and align with the official editing protocol in LiveEdit/MMEdit~\cite{chen2025lifelong,cheng2023edit}, including the same training/evaluation splits, and per-edit optimization settings.
For IKE, we use the MMEdit-aligned setup~\cite{cheng2023edit} to ensure a fair comparison under identical edit scopes.
For AlphaEdit~\cite{fangalphaedit}, we follow the official hyperparameters and construct the retain key set $K_0$ using E-VQA and E-IC train samples~\cite{cheng2023edit} to cover both modalities. Following Section~\ref{sec:edit_scoped_generalization}, we restrict editing to the last seven transformer layers, identified as the primary contributors.

We perform online updates with ScopeEdit using the closed-form solver (Eq.~\eqref{eq:prox_closed_form}), without iterative optimization. 
To make the setting more challenging, we use a per-edit batch size of 1 throughout.
We apply model-specific hyperparameters:
\begin{itemize}[noitemsep,nolistsep,leftmargin=*] 
\item \textbf{LLaVA-v1.5:} We set the low-rank write dimension to $r=512$ with $\alpha/r=2.0$. For the language backbone, we apply the dual-branch design with learning rates $\eta_{\mathrm{loc}}=0.1$ and $\eta_{\mathrm{sh}}=1.0$ for the local and shared branches, respectively. For the visual projection layer, we use a single low-rank write with $\eta_{\mathrm{vis}}=0.001$, without the local/shared branch decomposition, since applying dual-branch routing to the projection layer may distort low-level visual alignment. We update the last seven transformer layers and the visual projection layer. For each editable low-rank write, we initialize $A^{(l)}$ via \texttt{torch.nn.init.orthogonal} and $B^{(l)}$ as zero matrices, while freezing $A^{(l)}$ and updating only $B^{(l)}$.
\item \textbf{BLIP2-OPT:} We set the low-rank write dimension to $r=128$ with $\alpha/r=2.0$. For the language backbone, we use $\eta_{\mathrm{loc}}=0.05$ and $\eta_{\mathrm{sh}}=0.5$ for the local and shared branches, respectively. For the visual projection layer, we similarly use a single low-rank write with $\eta_{\mathrm{vis}}=0.001$ and do not apply branch decomposition. We update the last seven transformer layers and the visual projection layer, using the same initialization and freezing strategy as above. \end{itemize}

\subsubsection{Codebase}
For fair comparison, all methods are implemented within the unified \textsc{EasyEdit} framework~\cite{cheng2023edit} and evaluated on a single NVIDIA H20 (96GB) GPU.
% Our code is publicly available at \url{https://github.com/lab-klc/ScopeEdit}.

\begin{table*}[t]
\centering
\setlength{\tabcolsep}{2.8pt}
\renewcommand{\arraystretch}{1.02}
\caption{Online editing results on E-VQA and E-IC for BLIP2-OPT and LLaVA-v1.5 under different edit horizons. “Rel.”, “T/M-Gen.”, and “T/M-Loc.” abbreviate \emph{Reliability}, \emph{Generality}, and \emph{Locality} (for text/modal evaluations), respectively.
The subscript of each method (e.g., $_1$, $_{100}$) denotes the number of online edits performed. Rows shaded in light purple indicate \emph{parameter-modifying} methods.}
\label{tab:online_editing}
\resizebox{\textwidth}{!}{%
% \scriptsize
\begin{tabular}{>{\centering\arraybackslash}p{2.5em}| c |cccccc |cccccc}
\toprule
\multirow{2}{*}{\textbf{Model}} & \multirow{2}{*}{\textbf{Methods}} &
\multicolumn{6}{c|}{\textbf{E-VQA}} & \multicolumn{6}{c}{\textbf{E-IC}} \\
\cmidrule(lr){3-8} \cmidrule(lr){9-14}
& & Rel.$^\uparrow$ & T-Gen.$^\uparrow$ & M-Gen.$^\uparrow$ & T-Loc.$^\uparrow$ & M-Loc.$^\uparrow$ & Avg.$^\uparrow$ &
Rel.$^\uparrow$ & T-Gen.$^\uparrow$ & M-Gen.$^\uparrow$ & T-Loc.$^\uparrow$ & M-Loc.$^\uparrow$ & Avg.$^\uparrow$ \\
\midrule

% ===================== Model: BLIP2-OPT =====================
\multirow{18}{*}{\rotatebox[origin=c]{90}{\textbf{BLIP2-OPT (2.7B)}}}
% \multirow{16}{*}{\rotatebox[origin=c]{90}{\textbf{BLIP2-OPT (2.7B)}}}

& \cellcolor{lightgray}FT-L$_{1}$
  & \cellcolor{lightgray}100.00
  & \cellcolor{lightgray}100.00
  & \cellcolor{lightgray}60.00
  & \cellcolor{lightgray}94.74
  & \cellcolor{lightgray}100.00
  & \cellcolor{lightgray}90.95
  & \cellcolor{lightgray}96.77
  & \cellcolor{lightgray}95.02
  & \cellcolor{lightgray}90.72
  & \cellcolor{lightgray}90.05
  & \cellcolor{lightgray}68.27
  & \cellcolor{lightgray}88.16 \\

& \cellcolor{lightgray}FT-M$_{1}$
  & \cellcolor{lightgray}100.00
  & \cellcolor{lightgray}96.67
  & \cellcolor{lightgray}63.33
  & \cellcolor{lightgray}100.00
  & \cellcolor{lightgray}73.33
  & \cellcolor{lightgray}86.67
  & \cellcolor{lightgray}100.00
  & \cellcolor{lightgray}100.00
  & \cellcolor{lightgray}76.92
  & \cellcolor{lightgray}100.00
  & \cellcolor{lightgray}24.79
  & \cellcolor{lightgray}80.34 \\

& \cellcolor{lightgray}MEND$_{1}$
  & \cellcolor{lightgray}100.00
  & \cellcolor{lightgray}100.00
  & \cellcolor{lightgray}100.00
  & \cellcolor{lightgray}60.61
  & \cellcolor{lightgray}33.33
  & \cellcolor{lightgray}78.79
  & \cellcolor{lightgray}100.00
  & \cellcolor{lightgray}100.00
  & \cellcolor{lightgray}\textbf{100.00}
  & \cellcolor{lightgray}84.21
  & \cellcolor{lightgray}100.00
  & \cellcolor{lightgray}96.84 \\

& \cellcolor{lightgray}AlphaEdit$_{1}$
  & \cellcolor{lightgray}60.00
  & \cellcolor{lightgray}50.00
  & \cellcolor{lightgray}40.00
  & \cellcolor{lightgray}91.64
  & \cellcolor{lightgray}73.33
  & \cellcolor{lightgray}62.99
  & \cellcolor{lightgray}30.77
  & \cellcolor{lightgray}30.77
  & \cellcolor{lightgray}30.77
  & \cellcolor{lightgray}89.47
  & \cellcolor{lightgray}100.00
  & \cellcolor{lightgray}56.35 \\

& \cellcolor{lightgray}M-ORE$_{1}$
  & \cellcolor{lightgray}100.00 
  & \cellcolor{lightgray}100.00 
  & \cellcolor{lightgray}96.67 
  & \cellcolor{lightgray}100.00 
  & \cellcolor{lightgray}100.00
  & \cellcolor{lightgray}99.45
  & \cellcolor{lightgray}100.00 
  & \cellcolor{lightgray}100.00
  & \cellcolor{lightgray}84.62 
  & \cellcolor{lightgray}100.00 
  & \cellcolor{lightgray}100.00 
  & \cellcolor{lightgray}97.44 \\

& SERAC$_{1}$ & 100.00 & 100.00 & 100.00 & 89.47 & 80.00 & 93.89 & 97.38 & 95.52 & 86.11 & 100.00 & 63.82 & 88.57 \\
& IKE$_{1}$   & 100.00 & 100.00 & 100.00 & 52.63 & 13.33 & 73.19 & 100.00 & 100.00 & 100.00 & 63.16 & 10.00 & 74.63 \\
& LiveEdit$_{1}$  & 92.68 & 92.33 & 89.25 & 100.00 & 95.57 & 93.97 & 80.67 & 80.67 & 77.79 & 100.00 & 98.09 & 87.44 \\
  & \cellcolor{ours}\textbf{ScopeEdit$_{1}$}
  & \cellcolor{ours}\textbf{100.00} 
  & \cellcolor{ours}\textbf{100.00} 
  & \cellcolor{ours}\textbf{100.00}
  & \cellcolor{ours}\textbf{100.00} 
  & \cellcolor{ours}\textbf{100.00} 
  & \cellcolor{ours}\textbf{100.00}
  & \cellcolor{ours}\textbf{100.00} 
  & \cellcolor{ours}\textbf{100.00} 
  & \cellcolor{ours}92.31 
  & \cellcolor{ours}\textbf{100.00} 
  & \cellcolor{ours}\textbf{100.00} 
  & \cellcolor{ours}\textbf{98.72}\\

\cmidrule{2-14}

& \cellcolor{lightgray}FT-L$_{100}$
  & \cellcolor{lightgray}26.00
  & \cellcolor{lightgray}18.00
  & \cellcolor{lightgray}11.00
  & \cellcolor{lightgray}86.28
  & \cellcolor{lightgray}53.12
  & \cellcolor{lightgray}38.88
  & \cellcolor{lightgray}79.45
  & \cellcolor{lightgray}71.69
  & \cellcolor{lightgray}57.82
  & \cellcolor{lightgray}92.23
  & \cellcolor{lightgray}55.18
  & \cellcolor{lightgray}71.27 \\

& \cellcolor{lightgray}FT-M$_{100}$
  & \cellcolor{lightgray}58.40
  & \cellcolor{lightgray}50.85
  & \cellcolor{lightgray}43.69
  & \cellcolor{lightgray}\textbf{100.00}
  & \cellcolor{lightgray}46.85
  & \cellcolor{lightgray}59.96
  & \cellcolor{lightgray}67.18
  & \cellcolor{lightgray}62.17
  & \cellcolor{lightgray}51.63
  & \cellcolor{lightgray}\textbf{100.00}
  & \cellcolor{lightgray}9.81
  & \cellcolor{lightgray}58.16 \\

& \cellcolor{lightgray}MEND$_{100}$
  & \cellcolor{lightgray}1.00
  & \cellcolor{lightgray}1.00
  & \cellcolor{lightgray}1.00
  & \cellcolor{lightgray}90.42
  & \cellcolor{lightgray}77.20
  & \cellcolor{lightgray}34.12
  & \cellcolor{lightgray}0.00
  & \cellcolor{lightgray}0.00
  & \cellcolor{lightgray}0.00
  & \cellcolor{lightgray}46.96
  & \cellcolor{lightgray}54.05
  & \cellcolor{lightgray}20.20 \\

& \cellcolor{lightgray}AlphaEdit$_{100}$
  & \cellcolor{lightgray}36.83
  & \cellcolor{lightgray}28.50
  & \cellcolor{lightgray}26.78
  & \cellcolor{lightgray}86.35
  & \cellcolor{lightgray}69.97
  & \cellcolor{lightgray}49.69
  & \cellcolor{lightgray}27.94
  & \cellcolor{lightgray}28.57
  & \cellcolor{lightgray}26.69
  & \cellcolor{lightgray}90.52
  & \cellcolor{lightgray}78.83
  & \cellcolor{lightgray}50.51 \\
  
& \cellcolor{lightgray}M-ORE$_{100}$
  & \cellcolor{lightgray}99.13 
  & \cellcolor{lightgray}90.34
  & \cellcolor{lightgray}83.91 
  & \cellcolor{lightgray}99.29 
  & \cellcolor{lightgray}96.50 
  & \cellcolor{lightgray}93.77
  & \cellcolor{lightgray}90.14 
  & \cellcolor{lightgray}87.75 
  & \cellcolor{lightgray}66.31 
  & \cellcolor{lightgray}97.21 
  & \cellcolor{lightgray}94.97 
  & \cellcolor{lightgray}87.28 \\

& SERAC$_{100}$ & 88.03 & 85.18 & 88.03 & 89.95 & 37.60 & 77.76 & 74.02 & 63.79 & 56.93 & 77.52 & 42.94 & 63.04 \\
& IKE$_{100}$   & 83.53 & 83.00 & 84.72 & 74.63 & 6.37 & 66.45 & 81.18 & 71.13 & 84.30 & 79.90 & 11.93 & 65.69 \\
& LiveEdit$_{100}$  & 91.83 & 91.16 & 85.02 & 99.31 & 92.78 & 92.02 & 74.63 & 74.33 & 61.95 & 96.77 & 95.34 & 80.60 \\
  & \cellcolor{ours}\textbf{ScopeEdit$_{100}$}
  & \cellcolor{ours}\textbf{98.80} 
  & \cellcolor{ours}\textbf{91.84} 
  & \cellcolor{ours}\textbf{88.84}
  & \cellcolor{ours}98.94 
  & \cellcolor{ours}\textbf{96.80} 
  & \cellcolor{ours}\textbf{95.11}
  & \cellcolor{ours}\textbf{93.01} 
  & \cellcolor{ours}\textbf{91.75} 
  & \cellcolor{ours}\textbf{85.48} 
  & \cellcolor{ours}97.63 
  & \cellcolor{ours}\textbf{97.17} 
  & \cellcolor{ours}\textbf{93.01}\\
\midrule
\midrule

% ===================== Model: LLaVA-v1.5 =====================
% \multirow{16}{*}{\rotatebox[origin=c]{90}{\textbf{LLaVA-v1.5 (7B)}}}

% ===================== Model: LLaVA-v1.5 =====================
\multirow{18}{*}{\rotatebox[origin=c]{90}{\textbf{LLaVA-v1.5 (7B)}}}

& \cellcolor{lightgray}FT-L$_{1}$
  & \cellcolor{lightgray}95.66
  & \cellcolor{lightgray}99.00
  & \cellcolor{lightgray}81.84
  & \cellcolor{lightgray}89.38
  & \cellcolor{lightgray}89.98
  & \cellcolor{lightgray}91.17
  & \cellcolor{lightgray}100.00
  & \cellcolor{lightgray}100.00
  & \cellcolor{lightgray}89.80
  & \cellcolor{lightgray}91.67
  & \cellcolor{lightgray}28.01
  & \cellcolor{lightgray}81.89 \\

& \cellcolor{lightgray}FT-M$_{1}$
  & \cellcolor{lightgray}95.00
  & \cellcolor{lightgray}95.00
  & \cellcolor{lightgray}79.67
  & \cellcolor{lightgray}100.00
  & \cellcolor{lightgray}85.83
  & \cellcolor{lightgray}91.10
  & \cellcolor{lightgray}100.00
  & \cellcolor{lightgray}100.00
  & \cellcolor{lightgray}76.47
  & \cellcolor{lightgray}100.00
  & \cellcolor{lightgray}26.11
  & \cellcolor{lightgray}80.52 \\

& \cellcolor{lightgray}MEND$_{1}$
  & \cellcolor{lightgray}95.53
  & \cellcolor{lightgray}95.53
  & \cellcolor{lightgray}83.79
  & \cellcolor{lightgray}74.82
  & \cellcolor{lightgray}59.65
  & \cellcolor{lightgray}81.86
  & \cellcolor{lightgray}96.81
  & \cellcolor{lightgray}97.65
  & \cellcolor{lightgray}\textbf{96.44}
  & \cellcolor{lightgray}94.74
  & \cellcolor{lightgray}100.00
  & \cellcolor{lightgray}97.13 \\

& \cellcolor{lightgray}AlphaEdit$_{1}$
  & \cellcolor{lightgray}72.57
  & \cellcolor{lightgray}72.57
  & \cellcolor{lightgray}70.67
  & \cellcolor{lightgray}88.04
  & \cellcolor{lightgray}96.67
  & \cellcolor{lightgray}80.10
  & \cellcolor{lightgray}47.06
  & \cellcolor{lightgray}47.06
  & \cellcolor{lightgray}52.94
  & \cellcolor{lightgray}100.00
  & \cellcolor{lightgray}100.00
  & \cellcolor{lightgray}69.41 \\
  
  & \cellcolor{lightgray}M-ORE$_{1}$
  & \cellcolor{lightgray}100.00 & 
  \cellcolor{lightgray}100.00 
  & \cellcolor{lightgray}84.80 
  & \cellcolor{lightgray}100.00 & 
  \cellcolor{lightgray}100.00 & 
  \cellcolor{lightgray}97.47
  & \cellcolor{lightgray}100.00 & 
  \cellcolor{lightgray}100.00 & 
  \cellcolor{lightgray}88.24 & 
  \cellcolor{lightgray}100.00 & 
  \cellcolor{lightgray}100.00 & 
  \cellcolor{lightgray}98.04\\

& SERAC$_{1}$ & 90.00 & 90.00 & 60.00 & 100.00 & 23.33 & 72.67 & 88.61 & 90.17 & 81.45 & 98.24 & 50.83 & 81.86 \\
& IKE$_{1}$    & 50.00 & 50.00 & 50.00 & 58.33 & 15.00 & 44.67 & 94.12 & 94.12 & 94.12 & 62.50 & 12.50 & 71.47 \\
& LiveEdit$_{1}$  & 93.36 & 93.67 & 87.91 & 100.00 & 100.00 & 94.99 & 82.33 & 82.33 & 80.67 & 100.00 & 100.00 &  89.07 \\
  & \cellcolor{ours}\textbf{ScopeEdit$_{1}$}
  & \cellcolor{ours}\textbf{100.00} & \cellcolor{ours}\textbf{100.00} & \cellcolor{ours}\textbf{100.00} & \cellcolor{ours}\textbf{100.00} & \cellcolor{ours}\textbf{100.00} & \cellcolor{ours}\textbf{100.00}
  & \cellcolor{ours}\textbf{100.00} & \cellcolor{ours}\textbf{100.00} & 
  \cellcolor{ours}94.12 & 
  \cellcolor{ours}\textbf{100.00} & \cellcolor{ours}\textbf{100.00} & \cellcolor{ours}\textbf{99.02}\\

\cmidrule{2-14}

& \cellcolor{lightgray}FT-L$_{100}$
  & \cellcolor{lightgray}74.55
  & \cellcolor{lightgray}66.59
  & \cellcolor{lightgray}66.14
  & \cellcolor{lightgray}84.15
  & \cellcolor{lightgray}65.71
  & \cellcolor{lightgray}71.43
  & \cellcolor{lightgray}86.00
  & \cellcolor{lightgray}82.11
  & \cellcolor{lightgray}78.15
  & \cellcolor{lightgray}77.13
  & \cellcolor{lightgray}11.33
  & \cellcolor{lightgray}66.94 \\

& \cellcolor{lightgray}FT-M$_{100}$
  & \cellcolor{lightgray}85.67
  & \cellcolor{lightgray}81.75
  & \cellcolor{lightgray}67.03
  & \cellcolor{lightgray}\textbf{100.00}
  & \cellcolor{lightgray}46.19
  & \cellcolor{lightgray}76.13
  & \cellcolor{lightgray}84.57
  & \cellcolor{lightgray}80.05
  & \cellcolor{lightgray}62.88
  & \cellcolor{lightgray}\textbf{100.00}
  & \cellcolor{lightgray}8.32
  & \cellcolor{lightgray}67.16 \\

& \cellcolor{lightgray}MEND$_{100}$
  & \cellcolor{lightgray}1.09
  & \cellcolor{lightgray}1.09
  & \cellcolor{lightgray}1.01
  & \cellcolor{lightgray}75.33
  & \cellcolor{lightgray}61.67
  & \cellcolor{lightgray}28.04
  & \cellcolor{lightgray}0.11
  & \cellcolor{lightgray}0.08
  & \cellcolor{lightgray}0.04
  & \cellcolor{lightgray}25.52
  & \cellcolor{lightgray}31.33
  & \cellcolor{lightgray}11.42 \\

& \cellcolor{lightgray}AlphaEdit$_{100}$
  & \cellcolor{lightgray}71.33
  & \cellcolor{lightgray}70.93
  & \cellcolor{lightgray}70.67
  & \cellcolor{lightgray}88.45
  & \cellcolor{lightgray}77.67
  & \cellcolor{lightgray}75.81
  & \cellcolor{lightgray}51.98
  & \cellcolor{lightgray}48.21
  & \cellcolor{lightgray}47.23
  & \cellcolor{lightgray}92.59
  & \cellcolor{lightgray}56.90
  & \cellcolor{lightgray}59.38 \\

  & \cellcolor{lightgray}M-ORE$_{100}$
  & \cellcolor{lightgray}93.51
  & \cellcolor{lightgray}90.15
  & \cellcolor{lightgray}80.76
  & \cellcolor{lightgray}96.08
  & \cellcolor{lightgray}89.98
  & \cellcolor{lightgray}90.10
  & \cellcolor{lightgray}94.58
  & \cellcolor{lightgray}93.47
  & \cellcolor{lightgray}78.29
  & \cellcolor{lightgray}97.48
  & \cellcolor{lightgray}89.21
  & \cellcolor{lightgray}90.61 \\

& SERAC$_{100}$ & 87.71 & 85.03 & 67.13 & 92.27 & 20.83 & 70.59 & 73.38 & 71.00 & 59.72 & 72.88 & 25.85 & 60.57 \\
& IKE$_{100}$    & 38.87 & 35.85 & 39.40 & 46.22 & 11.24 & 34.32 & 78.78 & 77.37 & 78.07 & 53.86 & 12.88 & 60.19 \\
& LiveEdit$_{100}$  & 90.22 & 91.39 & 81.49 & 98.05 & \textbf{95.27} & 91.28 & 78.49 & 78.77 & 65.50 & \textbf{98.77} & \textbf{96.76} & 83.66 \\
& \cellcolor{ours}\textbf{ScopeEdit$_{100}$}
  & \cellcolor{ours}\textbf{92.52} & \cellcolor{ours}\textbf{96.05} & \cellcolor{ours}\textbf{84.18} 
  & \cellcolor{ours}98.69 
  & \cellcolor{ours}91.39 
  & \cellcolor{ours}\textbf{92.57}
  & \cellcolor{ours}\textbf{95.42} 
  & \cellcolor{ours}\textbf{94.66} 
  & \cellcolor{ours}\textbf{86.35} 
  & \cellcolor{ours}98.42 
  & \cellcolor{ours}92.26 
  & \cellcolor{ours}\textbf{93.42}\\

\bottomrule
\end{tabular}
}
\end{table*}

% ===== Table =====
\begin{table*}[h]
\centering
\setlength{\tabcolsep}{4pt}
\renewcommand{\arraystretch}{1.05}
\caption{
Ablation study of ScopeEdit design choices on the E-IC task for LLaVA-v1.5.
Shaded rows denote the full ScopeEdit model. Colored arrows in parentheses
indicate changes relative to ScopeEdit under the same edit horizon.
}
\label{tab:ablation_struct}
\resizebox{\textwidth}{!}{%
\begin{tabular}{l|c|c|c|c|c|c}
\toprule
\multirow{2}{*}{\textbf{Variant}} & \multicolumn{6}{c}{\textbf{E-IC}} \\
\cmidrule(lr){2-7}
& Rel.$^\uparrow$
& T-Gen.$^\uparrow$
& \textbf{M-Gen.}$^\uparrow$
& \textbf{T-Loc.}$^\uparrow$
& \textbf{M-Loc.}$^\uparrow$
& Avg.$^\uparrow$ \\
\midrule

\rowcolor{ours}
\textbf{ScopeEdit}$_{1}$              
  & $\textbf{100.00} \pm \textbf{0.00}$
  & $\textbf{100.00} \pm \textbf{0.00}$
  & $\textbf{92.16} \pm \textbf{3.39}$
  & $\textbf{100.00} \pm \textbf{0.00}$
  & $\textbf{100.00} \pm \textbf{0.00}$
  & $\textbf{98.43} \pm \textbf{0.68}$ \\

w/o Edit-Scoped Principle$_{1}$    
  & $100.00 \pm 0.00$
  & $100.00 \pm 0.00$
  & $90.20 \pm 3.39$\dneg{1.96}
  & $100.00 \pm 0.00$\dzero
  & $100.00 \pm 0.00$\dzero
  & $98.04 \pm 0.68$ \\

w/o Scope Separation$_{1}$           
  & $100.00 \pm 0.00$
  & $100.00 \pm 0.00$
  & $94.12 \pm 0.00$\dpos{1.96}
  & $97.22 \pm 4.81$\dneg{2.78}
  & $99.05 \pm 1.02$\dneg{0.95}
  & $98.08 \pm 1.17$ \\

w/o Scope Activation$_{1}$         
  & $100.00 \pm 0.00$
  & $100.00 \pm 0.00$ 
  & $92.16 \pm 3.39$\dzero
  & $95.83 \pm 0.00$\dneg{4.17}
  & $100.00 \pm 0.00$\dzero
  & $97.60 \pm 0.68$ \\

\cmidrule{1-7}

\rowcolor{ours}
\textbf{ScopeEdit}$_{100}$          
  & $\textbf{95.56} \pm \textbf{0.34}$
  & $\textbf{94.75} \pm \textbf{0.41}$
  & $\textbf{86.49} \pm \textbf{0.15}$
  & $\textbf{98.16} \pm \textbf{0.25}$
  & $\textbf{91.62} \pm \textbf{0.56}$
  & $\textbf{93.32} \pm \textbf{0.34}$ \\

w/o Edit-Scoped Principle$_{100}$  
  & $94.77 \pm 0.27$
  & $93.88 \pm 0.37$
  & $78.39 \pm 0.13$\dneg{8.10}
  & $97.64 \pm 0.18$\dneg{0.52}
  & $90.07 \pm 1.33$\dneg{1.55}
  & $90.95 \pm 0.46$ \\

w/o Scope Separation$_{100}$       
  & $93.71 \pm 0.56$ 
  & $93.06 \pm 0.40$
  & $89.55 \pm 1.19$\dpos{3.06}
  & $94.05 \pm 0.75$\dneg{4.11}
  & $87.72 \pm 1.92$\dneg{3.90}
  & $91.62 \pm 0.96$ \\

w/o Scope Activation$_{100}$       
  & $94.07 \pm 0.22$ 
  & $92.84 \pm 0.27$ 
  & $89.84 \pm 2.15$\dpos{3.35}
  & $94.98 \pm 1.11$\dneg{3.18}
  & $89.83 \pm 1.41$\dneg{1.79}
  & $92.31 \pm 1.03$ \\

\bottomrule
\end{tabular}
}
\end{table*}

\begin{figure*}[t]
    \centering
    \includegraphics[width=1\linewidth]{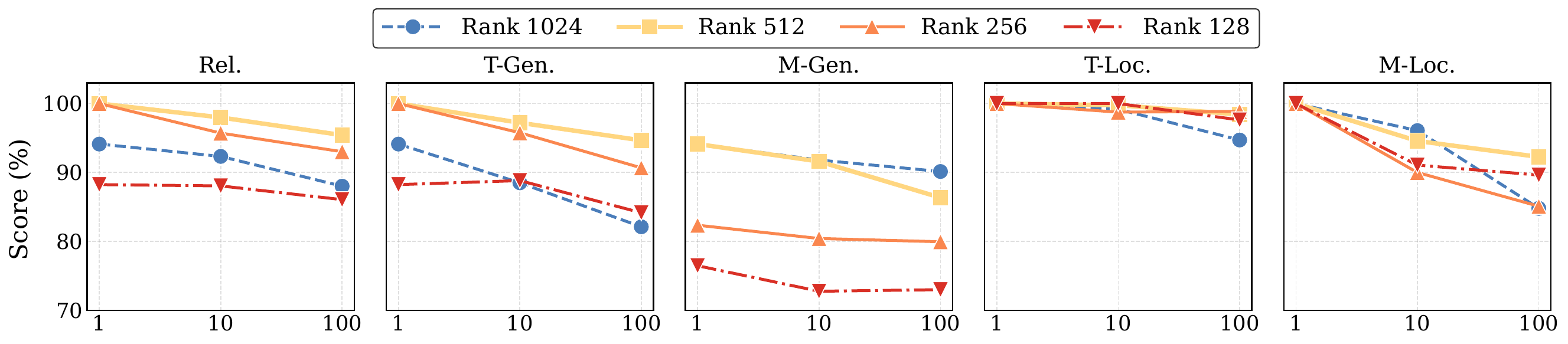}
    \caption{Sensitivity analysis of the low-rank write-interface dimension $r$ under different online edit horizons.}
    \label{fig:rank_sensitivity}
    \vspace{-3mm}
\end{figure*}

\subsection{Overall Online Editing Performance (RQ1)}
\label{sec:rq1}

\subsubsection{Performance on coarse-grained benchmarks}
\label{sec:q1}
Table~\ref{tab:online_editing} reports online editing results on E-VQA and E-IC across two MLLM backbones and different edit horizons. 
Existing editors exhibit different forms of scope-incorrect behavior. 
\emph{Parameter-modifying} methods such as FT and MEND can fit individual edits, but degrade sharply under long edit streams, suggesting accumulated interference and poor online stability. 
AlphaEdit sometimes preserves locality, but often fails to achieve reliable in-scope generalization under multimodal inputs. 
\emph{Parameter-preserving} methods such as SERAC, IKE, and LiveEdit better avoid direct weight drift, yet still show an imbalanced trade-off: they may preserve locality, but insufficiently propagate edits to in-scope cross-modal variants.

In contrast, \lsy{ScopeEdit} consistently achieves the best overall performance across all backbones, tasks, and edit horizons. 
Under $t{=}100$, it obtains the highest average scores in all four settings: $95.11$ on BLIP2-OPT E-VQA, $93.01$ on BLIP2-OPT E-IC, $92.57$ on LLaVA-v1.5 E-VQA, and $93.42$ on LLaVA-v1.5 E-IC. 
The improvements are especially clear on multimodal generalization, which reflects in-scope cross-modal generalization. 
On BLIP2-OPT E-IC at $t{=}100$, \lsy{ScopeEdit} improves M-Gen. from $66.31$ with M-ORE to $85.48$, while also increasing M-Loc. from $94.97$ to $97.17$. 
On LLaVA-v1.5 E-IC, it improves M-Gen. from $78.29$ to $86.35$ over M-ORE, while maintaining strong locality.
These results support our central claim: effective online MLLM editing should neither generalize indiscriminately nor suppress edit effects for locality. 
Instead, \emph{edits should generalize within their intended semantic scope while remaining constrained outside it}. 

\subsection{Design Analysis and Hyperparameter Sensitivity (RQ2)}
\label{sec:design_analysis}

\subsubsection{Ablation Study} Table~\ref{tab:ablation_struct} presents the ablation experiments demonstrating our core contributions, and our detailed setup is as follows:
\begin{itemize}[leftmargin=*, labelsep=0.8em, noitemsep,nolistsep]
    \item \textbf{w/o Edit-Scoped Principle} collapses \lsy{ScopeEdit} into a single-branch recursive editor, using one low-rank write space and one recursive statistic for all edits.

    \item \textbf{w/o Scope Separation} removes the orthogonal split between the modality-local absorption branch and the shared generalization branch, forcing both functions to share the same low-rank write space.

    \item \textbf{w/o Scope Activation} removes the evidence-based scope gate and always activates the shared generalization branch.
\end{itemize}

Overall, removing any component degrades the average score, and the gap becomes more pronounced under long edit streams, indicating that edit-scoped generalization is not achieved by stable recursive writing alone.

First, \emph{w/o Edit-Scoped Principle} collapses the two-branch formulation into a single recursive write path. 
While it remains reliable under $t{=}100$, its M-Gen. drops from $86.49$ to $78.39$. 
This shows that explicitly modeling editing as a scoped problem is necessary for in-scope cross-modal generalization beyond reliability and stability.

Second, \emph{w/o Scope Separation} obtains high M-Gen. but suffers clear locality degradation under long horizons, with T-Loc. and M-Loc. dropping from $98.16/91.62$ to $94.05/87.72$. 
This suggests that forcing local absorption and shared generalization into the same write space can over-expand the edit effect and harm out-of-scope locality preservation.

Third, \emph{w/o Scope Activation} encourages stronger cross-modal generalization by uniformly activating the shared branch, but this also over-generalizes the edit beyond its intended scope.
At $t{=}100$, T-Loc. and M-Loc. drop from $98.16/91.62$ to $94.98/89.83$, indicating increased out-of-scope leakage.
This confirms that cross-modal generalization must be gated by semantic evidence rather than applied uniformly.

\begin{table}[t]
\centering
\caption{Effect of different write-space basis initializations on E-IC with LLaVA-v1.5. We compare Gaussian, Xavier, data-driven,
and the proposed orthogonal initialization under short- and longhorizon online editing. The data-driven basis is initialized using
activation statistics collected from the training set.}
\label{tab:basis_init}
\setlength{\tabcolsep}{6.5pt}
\begin{tabular}{l|ccccc}
\toprule
\textbf{Basis Choice} & \textbf{Rel.} & \textbf{T-Gen.} & \textbf{M-Gen.} & \textbf{T-Loc.} & \textbf{M-Loc.} \\
\midrule
Gaussian$_1$          & 100.00 & 100.00 & 88.24 & 97.96 & 100.00 \\
Xavier$_1$            & 94.12 & 94.12 & 94.12 & 91.67 & 100.00  \\
Data-driven$_1$       & 100.00 & 100.00 & 94.12 & 91.67  & 100.00 \\[0.5ex]
\rowcolor{ours}
\textbf{Orthogonal}$_1$ & \textbf{100.00} & \textbf{100.00} & \textbf{94.12} & \textbf{100.00} & \textbf{100.00} \\
\midrule
Gaussian$_{100}$      & 83.19 & 80.83 & 72.26 & 94.87  & 85.44 \\
Xavier$_{100}$        & 11.39  & 11.39  & 11.39  & 96.64  & 56.83\\
Data-driven$_{100}$   & 94.23 & 88.31 & 79.55 & 92.36  & 90.83 \\[0.5ex]
\rowcolor{ours}
\textbf{Orthogonal}$_{100}$ & \textbf{95.42} & \textbf{94.66} & \textbf{86.35} & \textbf{98.42} & \textbf{92.26} \\
\bottomrule
\end{tabular}
\vspace{-2mm}
\end{table}

\begin{figure*}[t]
    \centering
    \captionsetup[subfloat]{font=scriptsize}

    \subfloat[MEND]{
        \includegraphics[width=0.325\textwidth]{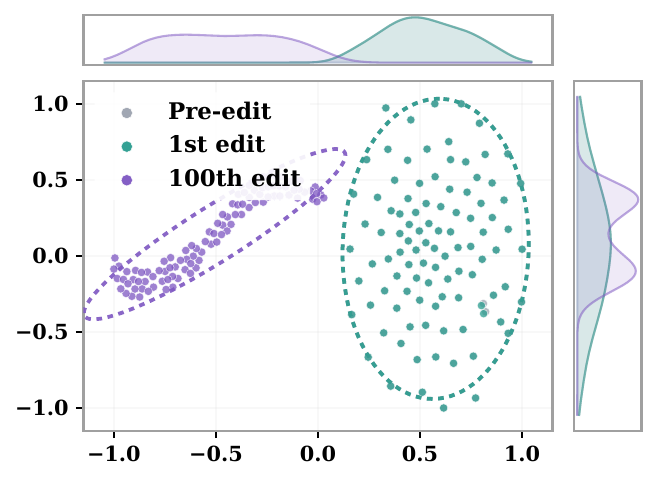}
        \label{fig:mend_drift_pre_scope}
    }
    \subfloat[M-ORE]{
        \includegraphics[width=0.325\textwidth]{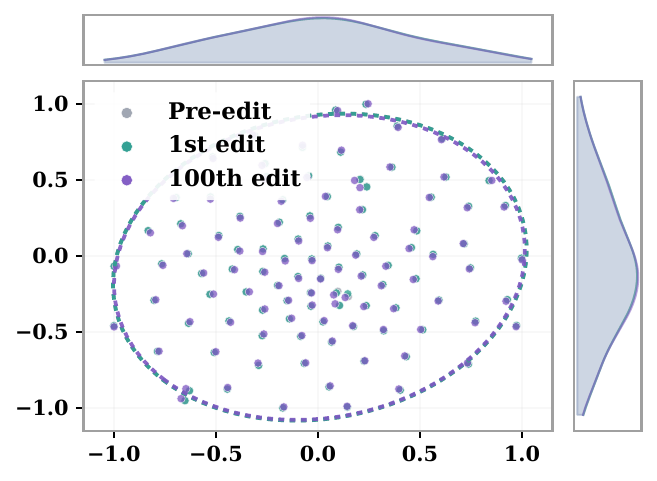}
        \label{fig:more_drift_pre_scope}
    }
    \subfloat[ScopeEdit]{
        \includegraphics[width=0.325\textwidth]{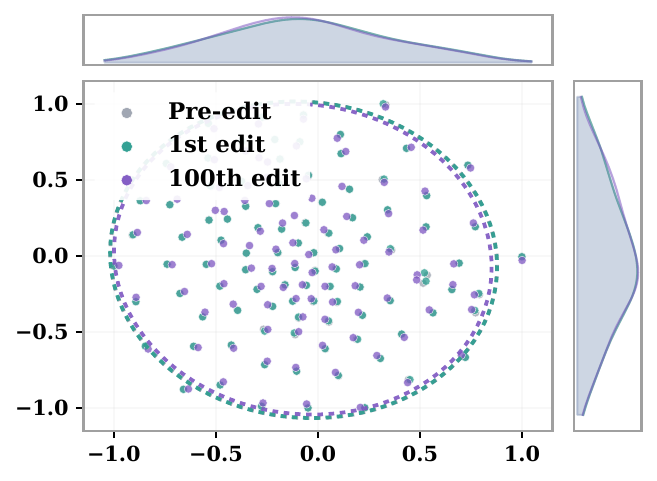}
        \label{fig:scopeedit_drift_pre_scope}
    }
    \caption{Sequential visualization of hidden-state distribution shifts under long-horizon online editing.
We compare MEND, M-ORE, and ScopeEdit on locality samples before editing, after the 1st edit, and after the 100th edit.
MEND exhibits clear representation drift after sequential edits, whereas M-ORE and ScopeEdit better preserve the pre-edit distribution.}
    \label{fig:drift_pre_1_100_scope}
    \vspace{-4mm}
\end{figure*}

\begin{figure*}[t]
    \centering
    \captionsetup[subfloat]{font=scriptsize}

    % ===================== Row 1: BLIP-2 =====================
    \subfloat{
        \includegraphics[width=0.24\textwidth]{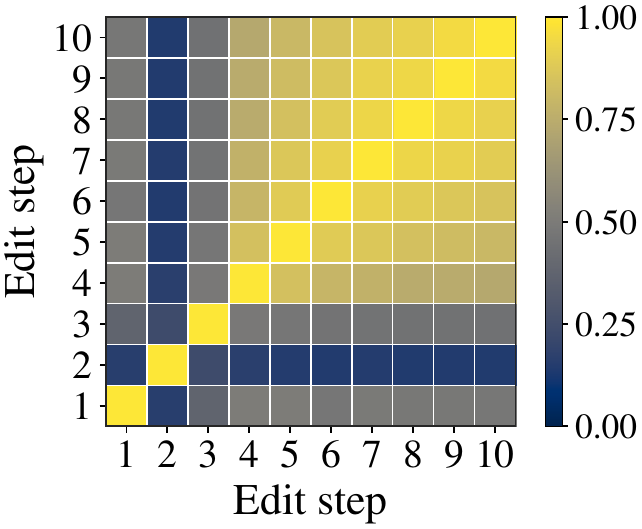}
        \label{fig:blip2-mend-overlap}
    }
    \subfloat{
        \includegraphics[width=0.24\textwidth]{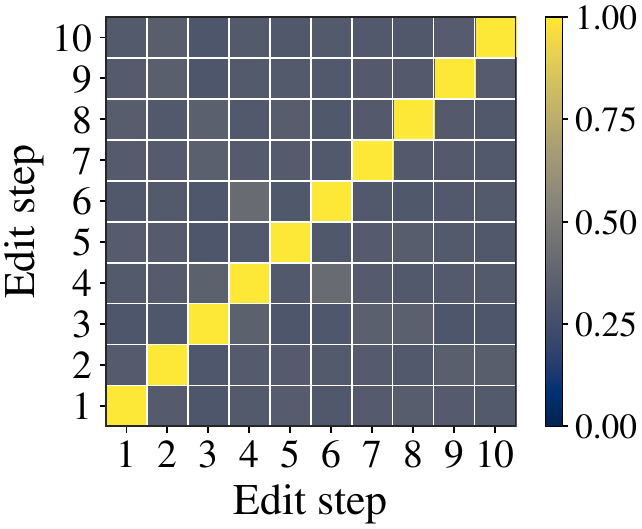}
        \label{fig:blip2-scopeedit-overlap}
    }
    \subfloat{
        \includegraphics[width=0.24\textwidth]{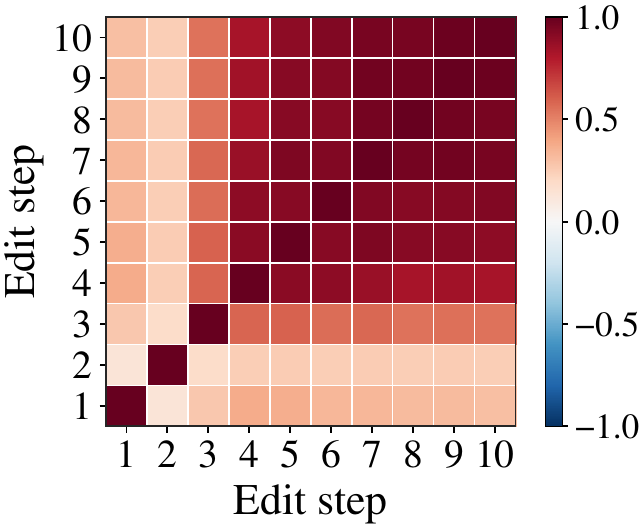}
        \label{fig:blip2-mend-cosine}
    }
    \subfloat{
        \includegraphics[width=0.24\textwidth]{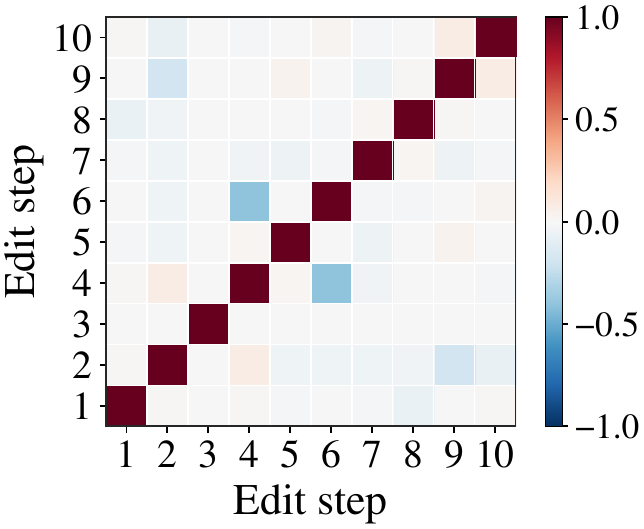}
        \label{fig:blip2-scopeedit-cosine}
    }

\vspace{-1.5mm}
\setcounter{subfigure}{0}

    % ===================== Row 2: LLaVA-v1.5 =====================
    \subfloat[MEND Top-$k$ coordinate overlap.]{
        \includegraphics[width=0.24\textwidth]{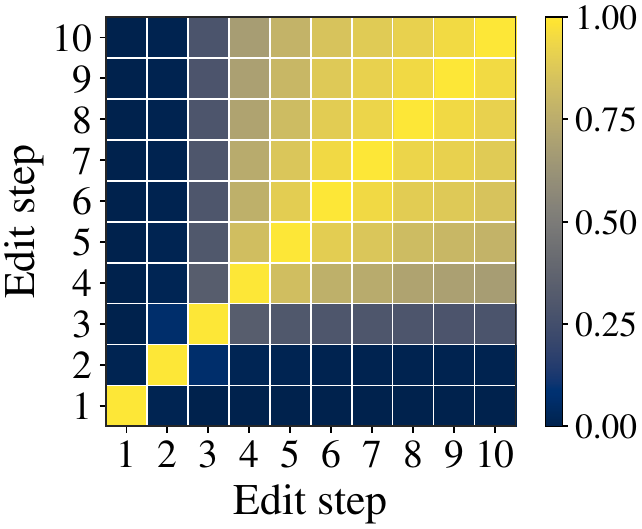}
        \label{fig:llava-mend-overlap}
    }
    \subfloat[ScopeEdit Top-$k$ coordinate overlap.]{
        \includegraphics[width=0.24\textwidth]{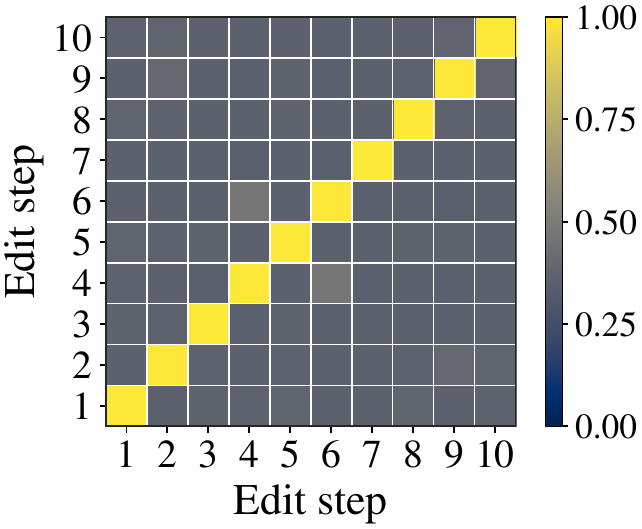}
        \label{fig:llava-scopeedit-overlap}
    }
    \subfloat[MEND weight cosine similarity.]{
        \includegraphics[width=0.24\textwidth]{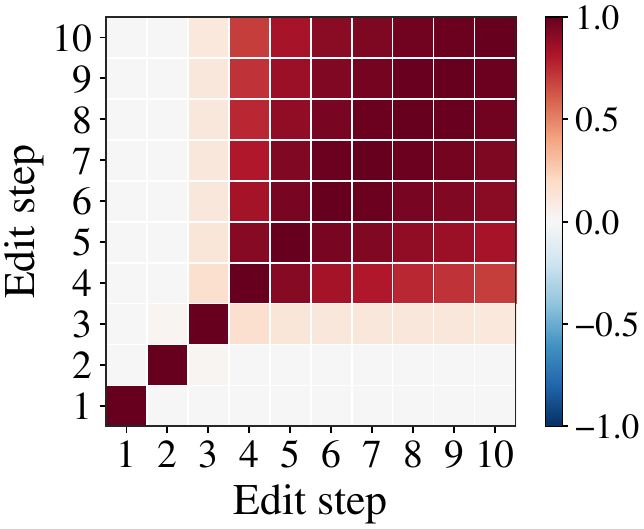}
        \label{fig:llava-mend-cosine}
    }
    \subfloat[ScopeEdit weight cosine similarity.]{
        \includegraphics[width=0.24\textwidth]{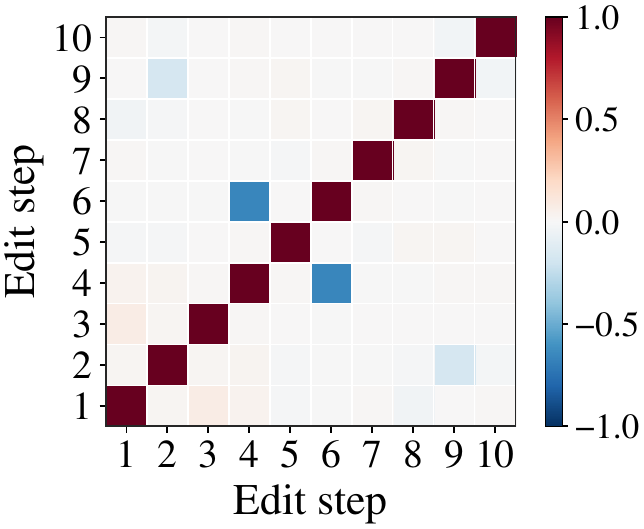}
        \label{fig:llava-scopeedit-cosine}
    }

    \caption{
        Pairwise weight-entanglement analysis across consecutive online edits.
The first row shows results on BLIP-2, and the second row shows results on LLaVA-v1.5.
For each backbone, we compare MEND and ScopeEdit using top-$k$ coordinate overlap and weight-update cosine similarity.
MEND exhibits high off-diagonal overlap and strong positive cosine similarity, indicating that sequential edits collapse into a shared edit core.
ScopeEdit maintains lower overlap and near-zero cosine similarity across edit pairs, suggesting more disentangled and stable online updates.
    }
    \label{fig:weight-entanglement-blip2-llava}
    \vspace{-3mm}
\end{figure*}

\subsubsection{Basis Initialization Analysis}
To examine the robustness of the orthogonal coordinate design more explicitly, we compare different initializations of the fixed basis $A^{(l)}$. As shown in Table~\ref{tab:basis_init}, random initializations such as Gaussian and Xavier are substantially less stable, especially under long edit horizons. Although the data-driven basis improves over random initialization, it still falls behind the orthogonal basis. In contrast, the proposed orthogonal initialization achieves the best overall performance at both $t$=1 and $t$=100, demonstrating that an orthonormal and fixed/drift-free coordinate system provides a more reliable low-rank write space for editing.

\subsubsection{Hyperparameter Analysis}
Figure~\ref{fig:rank_sensitivity} analyzes the effect of the low-rank write-interface dimension $r\in\{128,256,512,1024\}$. 
Overall, $r=512$ achieves the most balanced stability--plasticity trade-off. 
When the rank is too small, such as $r=128$, the restricted write capacity limits the absorption of edited knowledge, leading to weaker reliability and generality, especially on multimodal generalization. 
Increasing the rank provides a larger write space and generally improves edit expressiveness, but the gains are not strictly monotonic. 
Once the write space is sufficiently expressive, additional rank dimensions bring limited benefits and may introduce mild locality fluctuations under long edit streams due to increased editable degrees of freedom. 
These results suggest that ScopeEdit benefits from a write space that is expressive enough for editing but still sufficiently constrained for stable online updates. 
Therefore, we adopt $r=512$ as the default rank in our experiments.

\subsection{Long-Horizon Stability and Representation Shift (RQ3)}
\label{sec:rq3}

\subsubsection{Hidden representation stability}
We further investigate whether online editing induces accumulated representation drift on locality samples that are unrelated to the edited requests. 
Figure~\ref{fig:drift_pre_1_100_scope} compares three representative methods, i.e., MEND, M-ORE, and ScopeEdit. 
MEND causes a pronounced shift in the hidden-state distribution after sequential editing: the representations after the 100th edit clearly deviate from the pre-edit distribution, indicating that repeated parameter updates can gradually distort the model's representation space even for unrelated inputs. 
This observation explains the locality degradation and long-horizon instability of MEND, which is consistent with its performance collapse in Table~\ref{tab:online_editing}.

In contrast, M-ORE largely preserves the pre-edit distribution after both the 1st and 100th edits, suggesting that recursive history-aware preconditioning can effectively suppress accumulated drift. 
ScopeEdit further balances edit reliability, generality, locality, and long-horizon stability: its post-edit representation distributions remain highly overlapped with the pre-edit distribution, with no clear distributional separation even after 100 edits. 
This indicates that the proposed evidence-gated shared branch can enhance in-scope knowledge generalization while avoiding global perturbations to unrelated representations. 
Overall, these results demonstrate that ScopeEdit maintains stable edit-scoped behavior over long online edit streams.

\begin{figure*}[t]
    \centering
    \includegraphics[width=1\linewidth]{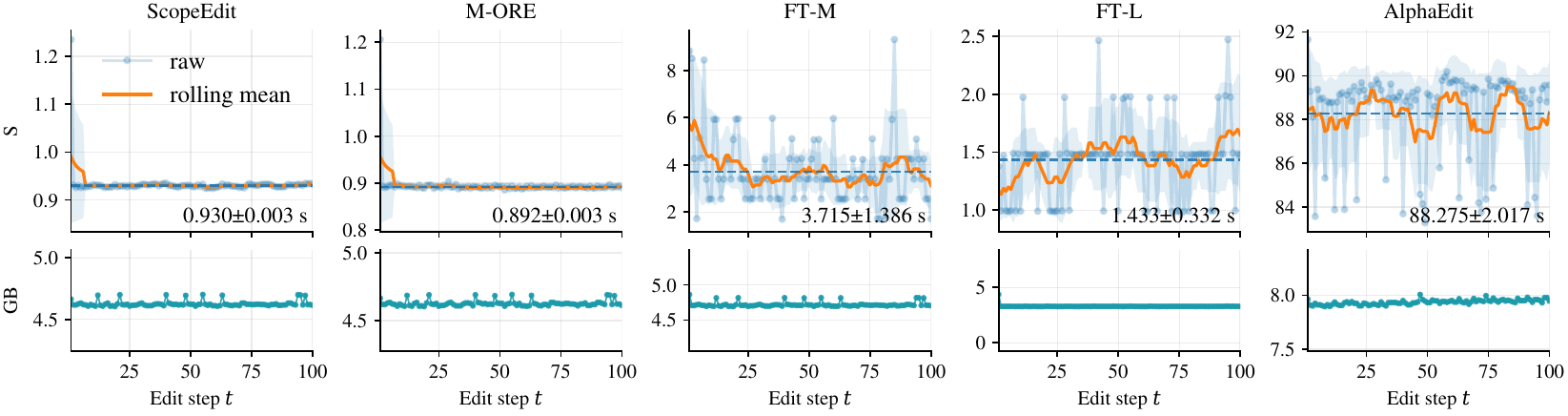}
    \caption{Efficiency metrics of editors on LLaVA-v1.5 over the online edit sequence $t$. 
\textbf{Top:} Edit latency (raw and rolling mean) . 
\textbf{Bottom:} Incremental peak memory ($\Delta$ peak mem).}
    \label{fig:scaling}
    \vspace{-3mm}
\end{figure*}

\subsubsection{Weight entanglement analysis}
We  analyze long-horizon stability from the perspective of parameter-update entanglement.
Figure~\ref{fig:weight-entanglement-blip2-llava} reports pairwise top-$k$ coordinate overlap and weight-update cosine similarity across consecutive edits.
For MEND, later edits increasingly reuse a compact set of update coordinates, leading to large off-diagonal top-$k$ overlaps.
Its weight-update cosine similarities also become strongly positive after several edits, indicating that different edits are repeatedly written into highly aligned parameter directions.
This suggests that sequential edits \emph{collapse into a shared edit core}, making newly injected edits interfere with previous ones and amplifying long-horizon drift.

In contrast, ScopeEdit shows substantially lower off-diagonal overlap and near-zero cosine similarity across most edit pairs on both backbones.
This indicates that its updates are less concentrated on the same coordinates and less directionally coupled across edit steps.
This observation aligns closely with the scope-separated write geometry and the recursive preconditioning mechanism: frequently exploited update directions are progressively suppressed, while local absorption and shared generalization are encoded into mutually independent orthogonal subspaces.
Therefore, ScopeEdit effectively reduces parameter-level entanglement, which explains its stronger stability under long online edit streams.

\begin{table}[t]
\footnotesize
\centering
\caption{
Per-edit computational and memory complexities under an online edit stream of length $t$.
Time/edit counts only editor-specific overhead, excluding the shared forward/backward cost for edit-gradient computation.
Here $r\ll d$, $|\mathcal{L}|$ denotes the number of edited layers, and $p_{\text{mem}}$ denotes the size of an edit-specific stored item.
}
\label{tab:complexity}
\setlength{\tabcolsep}{5pt}
\resizebox{\columnwidth}{!}{
\begin{tabular}{l|c|c|c}
\toprule
\textbf{Method} 
& \textbf{Time / edit} 
& \textbf{Memory} 
& $\boldsymbol{t}$\textbf{-dep.} \\
\midrule
\rowcolor{ours}
\textbf{ScopeEdit (ours)}
& $O\!\big(|\mathcal{L}|\,d_{\text{out}}r^2\big)$
& $O\!\big(|\mathcal{L}|(d_{\text{out}}r+r^2)\big)$
& $O(1)$ \\
\midrule
Locate-then-edit
& $O(d^3)$ or $O(d^2)$
& $O(d^2)$
& $O(1)$ \\
Null-space constraint
& $O(d\,t^2)$
& $O(d\,t)$
& $\uparrow$ \\
Parameter-preserving
& $O(t)$ or $O(\log t)$
& $O(t\,p_{\text{mem}})$
& $\uparrow$ \\
Naive finetuning
& $O\!\big(|\mathcal{L}|\,d_{\text{out}}r\big)$
& $O\!\big(|\mathcal{L}|\,d_{\text{out}}r\big)$
& $O(1)$ \\
\bottomrule
\end{tabular}
}
\vspace{-4mm}
\end{table}

\subsection{Online Editing Efficiency Analysis (RQ4)}
\label{sec:rq4}

We analyze the \emph{editor-specific} overhead of \lsy{ScopeEdit} under an online edit stream. 
Let $r=r_\ell+r_s\ll d$ denote the total low-rank edit dimension, where $r_\ell$ and $r_s$ are the ranks of the local absorption and shared generalization branches, respectively. 
Let $d$ and $d_{\mathrm{out}}$ be the FFN key/input and output dimensions, and let $\mathcal{L}$ be the set of edited layers. 
\lsy{ScopeEdit} performs each edit through fixed low-rank write coordinates and maintains only branch-wise recursive statistics in the edit subspace. 
Since \lsy{ScopeEdit} updates its branch-wise statistics using only the current-step keys
$
\mathcal{K}_t^{(l)}
=
\{k_t^{(l)}(x^q), \; k_t^{(l)}(x^v,x^q)\}
$
rather than storing past edit instances, the per-edit overhead is
$
\small
O\!\big(|\mathcal{L}|d_{\mathrm{out}}r^2\big),
$
and the online memory cost is
$
\small
O\!\Big(|\mathcal{L}|(d_{\mathrm{out}}r+r^2)\Big).
$
Both are independent of the edit stream length $t$. 
As summarized in Table~\ref{tab:complexity}, \lsy{ScopeEdit} avoids the dense second-order solvers required by locate-then-edit methods and does not store edit-specific memories or experts as parameter-preserving methods do. 
Thus, it provides constant per-edit compute and online memory under strict streaming constraints.

We further validate this analysis with empirical efficiency measurements in Figure~\ref{fig:scaling}.
\lsy{ScopeEdit} exhibits nearly flat per-edit latency across the edit stream, with an average time of $0.930\pm0.003$ seconds per edit.
This is comparable to M-ORE ($0.892\pm0.003$s), faster than FT-L ($1.433\pm0.332$s) and FT-M ($3.715\pm1.386$s), and substantially more efficient than AlphaEdit ($88.275\pm2.017$s), whose dense projection step dominates the runtime.
The memory curve of \lsy{ScopeEdit} also remains stable over time, showing no monotonic growth as $t$ increases.
Although naive finetuning is asymptotically cheaper, Table~\ref{tab:online_editing} shows that it sacrifices multimodal locality under sequential edits. 
\lsy{ScopeEdit} achieves a better quality--efficiency trade-off by maintaining constant online overhead while preserving both edit generalization and locality.

\begin{figure*}
    \centering
    \includegraphics[width=1\linewidth]{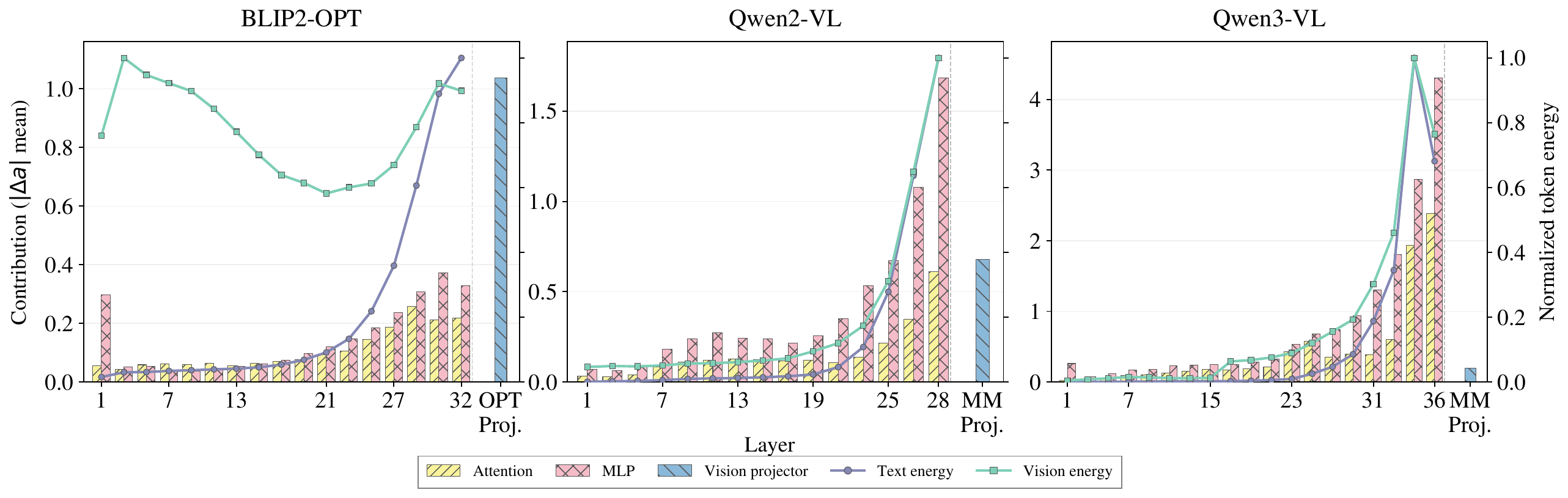}
    \caption{Cross-model validation of the layer-wise phenomenon. BLIP2-OPT, Qwen2-VL, and Qwen3-VL exhibit similar late-layer concentration patterns in module contributions and token energy, suggesting that the observed phenomenon is not model-specific.}
    \label{fig:combined_layer}
    \vspace{-2mm}
\end{figure*}

\begin{proof}[\textbf{Complexity Derivation}]
For each edited layer $l\in\mathcal{L}$, \lsy{ScopeEdit} maintains two branches $b\in\{\mathrm{loc},\mathrm{sh}\}$ with ranks $r_{\mathrm{loc}}=r_\ell$, $r_{\mathrm{sh}}=r_s$, and $r=r_\ell+r_s$. 
Each branch contains a fixed basis $A_b^{(l)}\in\mathbb{R}^{r_b\times d}$, an editable coefficient matrix $B_b^{(l)}\in\mathbb{R}^{d_{\mathrm{out}}\times r_b}$, and a recursive preconditioner $P_b^{(l)}\in\mathbb{R}^{r_b\times r_b}$.

At edit step $t$, projecting the branch-specific key,
$
z_{t,b}^{(l)}=A_b^{(l)}k_{t,b}^{(l)},
$
costs $O(r_b d)$. 
The history-preconditioned write
$
\Delta B_{t,b}^{(l)}
=
-\eta G_{t,b}^{(l)}P_{t-1,b}^{(l)}
$
with $G_{t,b}^{(l)}\in\mathbb{R}^{d_{\mathrm{out}}\times r_b}$ costs $O(d_{\mathrm{out}}r_b^2)$. 
The Sherman--Morrison update of $P_b^{(l)}$ in Eq.~\eqref{eq:SM_branch} is a rank-one correction in the $r_b$-dimensional edit subspace and costs $O(r_b^2)$. 
Thus, the per-layer cost is
\begin{equation}
\small
\sum_{b\in\{\mathrm{loc},\mathrm{sh}\}}
O\!\left(r_b d+d_{\mathrm{out}}r_b^2+r_b^2\right).
\end{equation}
Using $r_\ell+r_s=r$ and $r_\ell^2+r_s^2\leq r^2$, the total per-edit overhead over all edited layers is
\begin{equation}
\small
O\!\Big(
|\mathcal{L}|(rd+d_{\mathrm{out}}r^2+r^2)
\Big)
\approx
O\!\big(|\mathcal{L}|d_{\mathrm{out}}r^2\big),
\end{equation}
where the approximation reflects that the preconditioned low-rank write dominates in practice.

For online memory, \lsy{ScopeEdit} stores only the branch-wise coefficient matrices $B_b^{(l)}$ and recursive preconditioners $P_b^{(l)}$:
\begin{equation}
\small
\sum_{b\in\{\mathrm{loc},\mathrm{sh}\}}
O(d_{\mathrm{out}}r_b+r_b^2)
=
O(d_{\mathrm{out}}r+r_\ell^2+r_s^2)
\leq
O(d_{\mathrm{out}}r+r^2)
\end{equation}
per edited layer.  
Therefore, the total online memory cost is
\begin{equation}
\small
O\!\Big(
|\mathcal{L}|(d_{\mathrm{out}}r+r^2)
\Big).
\end{equation}
The fixed bases $A_b^{(l)}$ are initialized once with resident cost $O(|\mathcal{L}|rd)$ if counted, but they do not grow with the edit stream. 
Since \lsy{ScopeEdit} updates its statistics using only current-step keys and stores no past edit instances, both the per-edit computation and online memory are independent of the number of edits $t$.
\end{proof}

\subsection{Architecture and Scenario Extension (RQ5)}
\label{sec:cross_arch_extension}

\subsubsection{Consistency of pilot observations across MLLMs}

Our pilot study in Section~\ref{sec:edit_scoped_generalization} is conducted on LLaVA-v1.5, raising a natural question of whether the observed edit-scoped behaviors are architecture-specific or generally present across MLLMs. 
To examine this, we further evaluate representative MLLM backbones with different vision-language alignment designs, including BLIP2-OPT~\cite{li2023blip}, Qwen2-VL~\cite{wang2024qwen2vlenhancingvisionlanguagemodels}, and Qwen3-VL~\cite{bai2025qwen3vltechnicalreport}. 
For each backbone, we apply the same online editing protocol and analyze whether edit-related cross-modal responses concentrate in deeper semantic layers.

Figure~\ref{fig:combined_layer} reports the layer-wise module contributions and token-energy profiles across different MLLM backbones. 
The results show a consistent pattern: edit-related responses are not uniformly distributed, but become increasingly concentrated in deeper language layers. 
Although BLIP2-OPT, Qwen2-VL, and Qwen3-VL adopt different vision-language alignment designs, their text and vision token energies all exhibit clear late-layer amplification, and the dominant module contributions also shift toward high-level semantic layers. 
This indicates that the pilot observation on LLaVA-v1.5 is not architecture-specific. 
Instead, in-scope cross-modal generalization generally emerges where visual and textual evidence has been sufficiently fused into shared semantic representations. 
These findings support our design of applying edit-scoped control in deeper editable modules.

\subsubsection{Extension to Real-World Scenarios}

Beyond controlled editing benchmarks~\cite{cheng2023edit}, we further evaluate \lsy{ScopeEdit} on VLKEB~\cite{Huang2024vlkeb}, a real-world vision-language knowledge editing benchmark. 
VLKEB contains more natural visual-textual knowledge updates and further evaluates \emph{portability}, which measures whether the edited knowledge can be applied to related knowledge queries. 

Following VLKEB, for an edit request $e_t=(x_t^e,y_t^e)$ with $x_t^e=(x_t^v,x_t^q)$, let $\mathcal{P}^{\mathrm{port}}_t$ denote its portability set, where each $(x^p_t,y^p_t)\in\mathcal{P}^{\mathrm{port}}_t$ is a related visual-textual query-answer pair constructed from the multimodal knowledge graph associated with the edited entity. 
The portability score is defined as
\begin{equation}
\mathbb{E}_{(x^p_t,y^p_t)\sim\mathcal{P}^{\mathrm{port}}_t}
\big[
\mathbb{I}(f_{\theta_t}(x^p_t)=y^p_t)
\big].
\end{equation}
Thus, VLKEB evaluates not only reliability, generality, and locality, but also whether the edited model can transfer the corrected knowledge to related real-world contexts.

Different from the main online editing protocol in Section~\ref{sec:q1}, we follow the sliding continual editing protocol of VLKEB~\cite{Huang2024vlkeb}. 
We evaluate two representative MLLM backbones, BLIP2-OPT and LLaVA-v1.5, and compare \lsy{ScopeEdit} with parameter-modifying editors, including FT-L, FT-M, and M-ORE, as well as the memory-based editor SERAC.

\begin{table}[t]
\centering
\setlength{\tabcolsep}{2.8pt}
\renewcommand{\arraystretch}{1.02}
\caption{Online editing results on VLKEB for BLIP2-OPT and LLaVA-v1.5 under the sliding continual editing protocol. 
The subscript of each method denotes the number of consecutive edits in the sliding evaluation, e.g., $_1$ and $_{100}$ indicate evaluation after 1 and 100 sequential edits, respectively. 
``Port.'' denotes portability.}
\label{tab:online_editing_vlkeb}
\resizebox{\columnwidth}{!}{%
\begin{tabular}{>{\centering\arraybackslash}p{2.5em}| c |ccccccc}
\toprule
\multirow{2}{*}{\textbf{Model}} & \multirow{2}{*}{\textbf{Methods}} &
\multicolumn{7}{c}{\textbf{VLKEB}} \\
\cmidrule(lr){3-9}
& & Rel.$^\uparrow$ & T-Gen.$^\uparrow$ & M-Gen.$^\uparrow$ & T-Loc.$^\uparrow$ & M-Loc.$^\uparrow$ & Port.$^\uparrow$ & Avg.$^\uparrow$ \\
\midrule

\multirow{10}{*}{\rotatebox[origin=c]{90}{\textbf{BLIP2-OPT (2.7B)}}}
& \cellcolor{lightgray}FT-L$_{1}$ & \cellcolor{lightgray}{\textbf{99.95}} & \cellcolor{lightgray}{\textbf{99.23}} & \cellcolor{lightgray}{\textbf{100.00}} & \cellcolor{lightgray}{72.20} & \cellcolor{lightgray}{\textbf{20.18}} & \cellcolor{lightgray}{17.50} & \cellcolor{lightgray}{68.18} \\
& \cellcolor{lightgray}FT-M$_{1}$ & \cellcolor{lightgray}{99.53} & \cellcolor{lightgray}{96.89} & \cellcolor{lightgray}{99.27} & \cellcolor{lightgray}{\textbf{100.00}} & \cellcolor{lightgray}{5.87} & \cellcolor{lightgray}{27.90} & \cellcolor{lightgray}{71.58} \\
& \cellcolor{lightgray}M-ORE$_{1}$ & \cellcolor{lightgray}{98.70} & \cellcolor{lightgray}{91.41} & \cellcolor{lightgray}{91.26} & \cellcolor{lightgray}{81.87} & \cellcolor{lightgray}{12.23} & \cellcolor{lightgray}{\textbf{36.01}} & \cellcolor{lightgray}{68.58} \\
& SERAC$_{1}$ & 90.42  & 89.23  & 90.50 & 100.00 & 2.47 & 12.62 & 64.21 \\
& \cellcolor{ours}\textbf{ScopeEdit$_{1}$} & \cellcolor{ours}{98.35} & \cellcolor{ours}{96.95} & \cellcolor{ours}{93.02} & \cellcolor{ours}{82.91} & \cellcolor{ours}{13.48} & \cellcolor{ours}{34.38} & \cellcolor{ours}{\textbf{69.85}} \\
\cmidrule{2-9}
& \cellcolor{lightgray}FT-L$_{100}$ & \cellcolor{lightgray}{47.18} & \cellcolor{lightgray}{46.61} & \cellcolor{lightgray}{46.34} & \cellcolor{lightgray}{18.49} & \cellcolor{lightgray}{1.17} & \cellcolor{lightgray}{10.04} & \cellcolor{lightgray}{28.31} \\
& \cellcolor{lightgray}FT-M$_{100}$ & \cellcolor{lightgray}{30.80} & \cellcolor{lightgray}{29.69} & \cellcolor{lightgray}{30.67} & \cellcolor{lightgray}{100.00} & \cellcolor{lightgray}{1.23} & \cellcolor{lightgray}{23.59} & \cellcolor{lightgray}{36.00} \\
& \cellcolor{lightgray}M-ORE$_{100}$ & \cellcolor{lightgray}{73.03} & \cellcolor{lightgray}{\textbf{74.83}} & \cellcolor{lightgray}{70.33} & \cellcolor{lightgray}{81.44} & \cellcolor{lightgray}{9.15} & \cellcolor{lightgray}{35.24} & \cellcolor{lightgray}{57.34} \\
& SERAC$_{100}$ & \textbf{90.54} & 35.43 & \textbf{90.54} & \textbf{100.00} & 2.46 & 13.30 & 55.38 \\
& \cellcolor{ours}\textbf{ScopeEdit$_{100}$} & \cellcolor{ours}{80.64} & \cellcolor{ours}{72.64} & \cellcolor{ours}{73.09} & \cellcolor{ours}{81.47} & \cellcolor{ours}{\textbf{10.52}} & \cellcolor{ours}{\textbf{36.65}} & \cellcolor{ours}{\textbf{59.17}} \\

\midrule
\midrule

\multirow{10}{*}{\rotatebox[origin=c]{90}{\textbf{LLaVA-v1.5 (7B)}}}
& \cellcolor{lightgray}FT-L$_{1}$ & \cellcolor{lightgray}{99.85} & \cellcolor{lightgray}{\textbf{99.30}} & \cellcolor{lightgray}{\textbf{99.80}} & \cellcolor{lightgray}{86.96} & \cellcolor{lightgray}{30.35} & \cellcolor{lightgray}{30.70} & \cellcolor{lightgray}{74.49} \\
& \cellcolor{lightgray}FT-M$_{1}$ & \cellcolor{lightgray}{100.00} & \cellcolor{lightgray}{98.98} & \cellcolor{lightgray}{98.75} & \cellcolor{lightgray}{\textbf{100.00}} & \cellcolor{lightgray}{19.87} & \cellcolor{lightgray}{55.48} & \cellcolor{lightgray}{78.85} \\
& \cellcolor{lightgray}M-ORE$_{1}$ & \cellcolor{lightgray}{94.57} & \cellcolor{lightgray}{87.44} & \cellcolor{lightgray}{91.31} & \cellcolor{lightgray}{87.41} & \cellcolor{lightgray}{\textbf{57.33}} & \cellcolor{lightgray}{\textbf{60.79}} & \cellcolor{lightgray}{79.81} \\
& SERAC$_{1}$ & 98.56 & 97.46 & 98.56 & 99.96 & 1.94 & 41.46 & 72.99 \\
& \cellcolor{ours}\textbf{ScopeEdit$_{1}$} & \cellcolor{ours}{\textbf{100.00}} & \cellcolor{ours}{93.37} & \cellcolor{ours}{99.10} & \cellcolor{ours}{87.06} & \cellcolor{ours}{53.05} & \cellcolor{ours}{60.42} & \cellcolor{ours}{\textbf{82.17}} \\
\cmidrule{2-9}
& \cellcolor{lightgray}FT-L$_{100}$ & \cellcolor{lightgray}{72.27} & \cellcolor{lightgray}{65.90} & \cellcolor{lightgray}{68.28} & \cellcolor{lightgray}{56.97} & \cellcolor{lightgray}{11.80} & \cellcolor{lightgray}{28.85} & \cellcolor{lightgray}{50.68} \\
& \cellcolor{lightgray}FT-M$_{100}$ & \cellcolor{lightgray}{69.58} & \cellcolor{lightgray}{67.28} & \cellcolor{lightgray}{66.68} & \cellcolor{lightgray}{\textbf{100.00}} & \cellcolor{lightgray}{2.53} & \cellcolor{lightgray}{45.93} & \cellcolor{lightgray}{58.67} \\
& \cellcolor{lightgray}M-ORE$_{100}$ & \cellcolor{lightgray}{93.31} & \cellcolor{lightgray}{87.21} & \cellcolor{lightgray}{90.47} & \cellcolor{lightgray}{84.36} & \cellcolor{lightgray}{\textbf{51.53}} & \cellcolor{lightgray}{60.62} & \cellcolor{lightgray}{77.92} \\
& SERAC$_{100}$ & \textbf{98.56} & 70.32 & \textbf{98.56} & 99.96 & 1.91 & 41.08 & 68.40 \\
& \cellcolor{ours}\textbf{ScopeEdit$_{100}$} & \cellcolor{ours}{94.47} & \cellcolor{ours}{\textbf{87.26}} & \cellcolor{ours}{91.40} & \cellcolor{ours}{84.17} & \cellcolor{ours}{51.24} & \cellcolor{ours}{\textbf{60.69}} & \cellcolor{ours}{\textbf{78.20}} \\
\bottomrule
\end{tabular}
}
\vspace{-3mm}
\end{table}

Table~\ref{tab:online_editing_vlkeb} shows that \lsy{ScopeEdit} remains effective and preserves stable edit-scoped behavior under real-world sliding continual editing.
First, fine-tuning-based methods achieve high scores at $t{=}1$ but degrade sharply after $100$ edits, revealing poor long-horizon stability in realistic visual-textual streams.
Second, recursive parameter-modifying editors, including M-ORE and \lsy{ScopeEdit}, substantially outperform traditional editors on portability, suggesting that stable parameter updates are more effective than direct fine-tuning or memory-based correction for transferring edited knowledge to related visual-textual contexts.
Third, compared with M-ORE, \lsy{ScopeEdit} achieves a better long-horizon trade-off across all evaluation dimensions.
At $t{=}100$, \lsy{ScopeEdit} achieves the best Avg. scores on both BLIP2-OPT and LLaVA-v1.5, reaching $59.17$ and $78.20$, respectively.
This indicates that \lsy{ScopeEdit} does not merely excel on isolated metrics, but provides a strong overall trade-off among reliability, generality, locality, and portability.

\begin{table}[t]
\centering
\footnotesize
\setlength{\tabcolsep}{2.8pt}
\caption{Extension to complex vision-language architectures on E-VQA and E-IC with LLaVA-v1.5 under short- and long-horizon online edits.}
\label{tab:extension_evqa_eic}
\begin{tabular}{l | c c c c c}
\toprule
Variant & Rel.$^\uparrow$ & T-Gen.$^\uparrow$ & M-Gen.$^\uparrow$ & T-Loc.$^\uparrow$ & M-Loc.$^\uparrow$  \\
\midrule
\multicolumn{6}{l}{\textit{E-VQA Task}} \\
\cellcolor{ours}\textbf{ScopeEdit$_1$ (default)}
& \cellcolor{ours}\textbf{100.00}
& \cellcolor{ours}\textbf{100.00}
& \cellcolor{ours}\textbf{100.00}
& \cellcolor{ours}\textbf{100.00}
& \cellcolor{ours}\textbf{100.00}\\
ScopeEdit w/ vis$_1$      & 63.33 & 60.00 & 51.33 & 100.00 & 100.00 \\
ScopeEdit w/ text$_1$      &100.00 & 100.00 & 50.00 & 100.00 & 100.00 \\
ScopeEdit w/ vis-text$_1$ & 100.00 & 100.00 & 100.00 & 100.00 & 100.00 \\
\cmidrule(lr){1-6}
\cellcolor{ours}\textbf{ScopeEdit$_{100}$ (default)}
& \cellcolor{ours}92.52
& \cellcolor{ours}\textbf{96.05}
& \cellcolor{ours}84.18
& \cellcolor{ours}98.69
& \cellcolor{ours}91.39 \\
ScopeEdit w/ vis$_{100}$      & 89.07 & 84.70 & 66.31 & \textbf{100.00} & 90.47 \\
ScopeEdit w/ text$_{100}$      &92.51 & 86.65 & 81.39 & 92.02 & 82.47 \\
ScopeEdit w/ vis-text$_{100}$ & \textbf{100.00} & 96.00 & \textbf{87.55} & 98.64 & \textbf{93.08} \\
\midrule
\midrule
\multicolumn{6}{l}{\textit{E-IC Task}} \\
\cellcolor{ours}\textbf{ScopeEdit$_1$ (default)}
& \cellcolor{ours}\textbf{100.00}
& \cellcolor{ours}\textbf{100.00}
& \cellcolor{ours}\textbf{94.12}
& \cellcolor{ours}\textbf{100.00}
& \cellcolor{ours}\textbf{100.00} \\
ScopeEdit w/ vis$_1$     & 58.82 & 52.94 & 52.94 & 100.00 & 100.00 \\
ScopeEdit w/ text$_1$    &100.00 & 100.00 & 94.12 & 95.83 & 100.00 \\
ScopeEdit w/ vis-text$_1$ & 100.00 & 100.00 & 94.12 & 100.00 & 100.00 \\
\cmidrule(lr){1-6}
\cellcolor{ours}\textbf{ScopeEdit$_{100}$ (default)}
& \cellcolor{ours}95.42
& \cellcolor{ours}\textbf{94.66}
& \cellcolor{ours}86.35
& \cellcolor{ours}98.42
& \cellcolor{ours}92.26 \\
ScopeEdit w/ vis$_{100}$     & 88.88 & 84.55 & 65.35 & \textbf{100.00} & 87.30 \\
ScopeEdit w/ text$_{100}$     &95.09 & 90.97 & 81.79 & 94.73 & \textbf{93.52} \\
ScopeEdit w/ vis-text$_{100}$ & \textbf{96.91} & 90.09 & \textbf{90.95} & 94.07 & 91.74 \\
\bottomrule
\end{tabular}
\vspace{-3mm}
\end{table}

\subsubsection{Extension to complex vision-language architectures}

Since the \emph{Edit-Scoped Online Editing Principle} is module-wise, each editable module can maintain its own scope-separated write geometry, branch-wise historical statistic $C^{(l)}$, and recursive preconditioner $P^{(l)}$. 
This allows modality-local absorption and shared cross-modal generalization to be applied not only to the projector and language layers, but also to vision-side modules.

To examine this extensibility, we evaluate \lsy{ScopeEdit} on LLaVA-v1.5 under both short- and long-horizon online editing settings. 
Besides the default setting that edits the projector and text-LLM modules, we consider three variants:
\begin{itemize}[leftmargin=*, labelsep=0.8em, noitemsep,nolistsep]
    \item \textbf{ScopeEdit w/ vis}: edits only the vision encoder and projector, testing whether vision-side modules alone can absorb multimodal corrections.
    \item \textbf{ScopeEdit w/ text}: edits only the text-LLM modules, testing whether language-side updates alone can absorb multimodal corrections.
    \item \textbf{ScopeEdit w/ vis-text}: jointly edits the vision encoder, projector, and text-LLM modules, testing the effectiveness of \lsy{ScopeEdit} under a more complex editable architecture.
\end{itemize}

For vision-side editing, the branch-specific projected keys are instantiated as
$
z_{t,\mathrm{loc}}^{(l)}=z_{\mathrm{loc}}^{(l)}(x_t^v)
$
for the local branch and
$
z_{t,\mathrm{sh}}^{(l)}=z_{\mathrm{sh}}^{(l)}(x_t^v,x_t^q)
$
for the shared branch. 
We edit only the last three layers of the vision encoder to avoid excessive perturbation to low-level visual representations.

Table~\ref{tab:extension_evqa_eic} evaluates the extension of \lsy{ScopeEdit} to more complex editable architectures. 
First, the default projector-plus-text-LLM setting already achieves a strong balance between edit absorption and locality preservation. 
Second, among the extended variants, \textbf{\lsy{ScopeEdit} w/ vis} preserves locality but substantially weakens reliability and generality, while \textbf{\lsy{ScopeEdit} w/ text} maintains strong text-side performance but provides limited multimodal generalization. 

Third, \textbf{\lsy{ScopeEdit} w/ vis-text} yields the strongest long-horizon M-Gen., improving it from $84.18$ to $87.55$ on E-VQA and from $86.35$ to $90.95$ on E-IC at $t{=}100$. 
These results suggest that incorporating vision-side modules can further enhance in-scope cross-modal generalization when combined with language-side editing, and the module-wise recursive design keeps locality competitive.

Beyond module-level extension within LLaVA-v1.5, we further test whether \lsy{ScopeEdit} can generalize to recent MLLM backbones with different vision-language architectures. 
We evaluate Qwen2-VL and Qwen3-VL on E-IC under the same online editing protocol and compare \lsy{ScopeEdit} with M-ORE across $t\in\{1,10,100\}$ edit horizons, using the same hyperparameters as LLaVA-v1.5 without architecture-specific tuning.

\begin{figure*}
    \centering
    \includegraphics[width=1\linewidth]{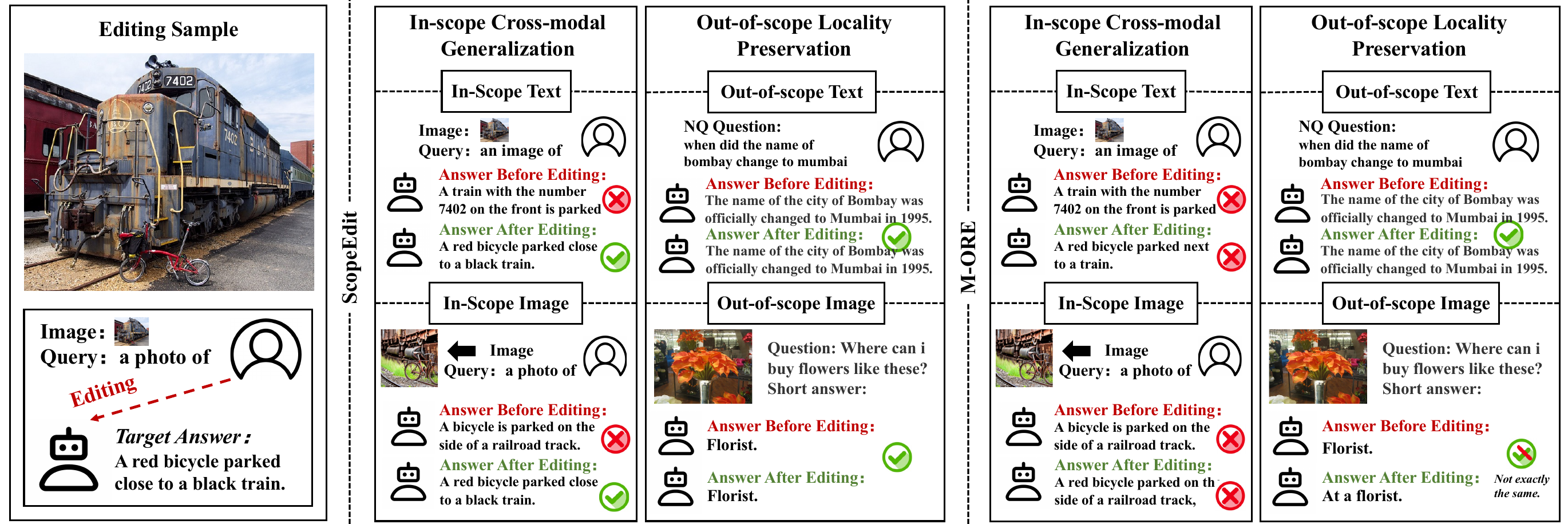}
    \caption{Case study of edit-scope control. Given the same multimodal edit, ScopeEdit correctly absorbs the target correction and generalizes it to in-scope text and image variants, while preserving unrelated textual and multimodal locality. In contrast, M-ORE either fails to propagate the edit to in-scope variants or leaks the edited knowledge to unrelated multimodal inputs.}
    \label{fig:case_study}
    \vspace{-3mm}
\end{figure*}

\begin{table}[t]
\centering
\setlength{\tabcolsep}{2.8pt}
\renewcommand{\arraystretch}{1.02}
\caption{Cross-architecture online editing results on E-IC with Qwen2-VL and Qwen3-VL under different edit horizons.}
\label{tab:qwen_online_editing}
\resizebox{\columnwidth}{!}{%
\begin{tabular}{>{\centering\arraybackslash}p{2.5em}|c|ccccc}
\toprule
\multirow{2}{*}{\textbf{Model}} & \multirow{2}{*}{\textbf{Methods}} &
\multicolumn{5}{c}{\textbf{E-IC}} \\
\cmidrule(lr){3-7}
& & Rel.$^\uparrow$ & T-Gen.$^\uparrow$ & M-Gen.$^\uparrow$ & T-Loc.$^\uparrow$ & M-Loc.$^\uparrow$ \\
\midrule

\multirow{6}{*}{\rotatebox[origin=c]{90}{\textbf{Qwen2-VL}}}
& \cellcolor{lightgray}M-ORE$_1$
& \cellcolor{lightgray}100.00 & \cellcolor{lightgray}100.00 & \cellcolor{lightgray}90.82 & \cellcolor{lightgray}100.00 & \cellcolor{lightgray}100.00 \\
& \cellcolor{ours}\textbf{ScopeEdit$_1$}
& \cellcolor{ours}\textbf{100.00} & \cellcolor{ours}\textbf{100.00} & \cellcolor{ours}\textbf{100.00}& \cellcolor{ours}\textbf{100.00}& \cellcolor{ours}\textbf{100.00} \\
\cmidrule{2-7}
& \cellcolor{lightgray}M-ORE$_{10}$
& \cellcolor{lightgray}\textbf{96.01} & \cellcolor{lightgray}\textbf{95.18} & \cellcolor{lightgray}85.85 & \cellcolor{lightgray}97.52 & \cellcolor{lightgray}95.00 \\
& \cellcolor{ours}\textbf{ScopeEdit$_{10}$}
& \cellcolor{ours}95.10 & \cellcolor{ours}92.50 & \cellcolor{ours}\textbf{90.75} & \cellcolor{ours}\textbf{97.83} & \cellcolor{ours}\textbf{96.67} \\
\cmidrule{2-7}
& \cellcolor{lightgray}M-ORE$_{100}$
& \cellcolor{lightgray}91.73 & \cellcolor{lightgray}91.01 & \cellcolor{lightgray}72.98 & \cellcolor{lightgray}87.42 & \cellcolor{lightgray}\textbf{88.92} \\
& \cellcolor{ours}\textbf{ScopeEdit$_{100}$}
& \cellcolor{ours}\textbf{92.11} & \cellcolor{ours}\textbf{91.40} & \cellcolor{ours}\textbf{77.96} & \cellcolor{ours}\textbf{89.37} & \cellcolor{ours}86.11 \\

\midrule

\multirow{6}{*}{\rotatebox[origin=c]{90}{\textbf{Qwen3-VL}}}
& \cellcolor{lightgray}M-ORE$_1$
& \cellcolor{lightgray}100.00 & \cellcolor{lightgray}100.00 & \cellcolor{lightgray}92.31 & \cellcolor{lightgray}100.00 & \cellcolor{lightgray}100.00 \\
& \cellcolor{ours}\textbf{ScopeEdit$_1$}
& \cellcolor{ours}\textbf{100.00} & \cellcolor{ours}\textbf{100.00} & \cellcolor{ours}\textbf{92.31} & \cellcolor{ours}\textbf{100.00} & \cellcolor{ours}\textbf{100.00} \\
\cmidrule{2-7}
& \cellcolor{lightgray}M-ORE$_{10}$
& \cellcolor{lightgray}94.20 & \cellcolor{lightgray}\textbf{93.43} & \cellcolor{lightgray}82.57 & \cellcolor{lightgray}91.28 & \cellcolor{lightgray}95.00 \\
& \cellcolor{ours}\textbf{ScopeEdit$_{10}$}
& \cellcolor{ours}\textbf{94.97} & \cellcolor{ours}91.68 & \cellcolor{ours}\textbf{84.80} & \cellcolor{ours}\textbf{94.97} & \cellcolor{ours}\textbf{100.00} \\
\cmidrule{2-7}
& \cellcolor{lightgray}M-ORE$_{100}$
& \cellcolor{lightgray}\textbf{93.89} & \cellcolor{lightgray}\textbf{91.80} & \cellcolor{lightgray}65.38 & \cellcolor{lightgray}89.70 & \cellcolor{lightgray}87.78 \\
& \cellcolor{ours}\textbf{ScopeEdit$_{100}$}
& \cellcolor{ours}91.16 & \cellcolor{ours}86.98 & \cellcolor{ours}\textbf{73.75} & \cellcolor{ours}\textbf{91.64} & \cellcolor{ours}\textbf{89.30} \\

\bottomrule
\end{tabular}
}
\vspace{-3mm}
\end{table}

Table~\ref{tab:qwen_online_editing} reports the results, showing that \lsy{ScopeEdit} remains effective on both Qwen2-VL and Qwen3-VL.
First, it consistently improves long-horizon multimodal generalization over M-ORE, increasing M-Gen. from $72.98$ to $77.96$ on Qwen2-VL and from $65.38$ to $73.75$ on Qwen3-VL at $t{=}100$. 
Second, \lsy{ScopeEdit} maintains competitive reliability and text generality, with only mild trade-offs on Qwen3-VL under long edit streams. 
Third, locality remains stable across horizons, especially on text-side locality. 
These results complement the LLaVA-v1.5 module-extension study, showing that \lsy{ScopeEdit} is not architecture-specific but provides a general edit-scoped mechanism for improving cross-modal generalization and preserving locality across other recent MLLM backbones.

\subsection{Case Study}

Figure~\ref{fig:case_study} provides an intuitive comparison between ScopeEdit and M-ORE on the same editing sample. ScopeEdit updates the target knowledge from ``train'' to ``bicycle'' and consistently transfers this correction to both text-side and image-side in-scope variants. Meanwhile, it preserves the outputs on unrelated textual and multimodal locality samples, indicating that the edit is confined to the intended scope. In contrast, M-ORE shows less precise scope control: the correction is not reliably activated for in-scope variants, while an unrelated multimodal sample is incorrectly affected by the edit. This case supports our motivation that online multimodal editing requires not only reliable knowledge injection, but also explicit scope-aware separation between in-scope generalization and out-of-scope preservation.

\section{Conclusion}
\label{sec:conclusion}

In this work, we studied online multimodal knowledge editing from the perspective of \emph{edit-scoped generalization}. 
We showed that reliable edits in MLLMs may still fail to generalize to semantically valid cross-modal variants or leak to unrelated inputs, indicating that reliability and long-horizon stability alone are insufficient for practical online editing. 
To address this issue, we proposed \lsy{ScopeEdit}, a scope-aware online editor that decomposes each update into a modality-local absorption branch and an evidence-gated shared generalization branch. 
By combining scope-separated low-rank write geometries, cross-modal evidence gating, and branch-wise recursive preconditioners, \lsy{ScopeEdit} enables stable online editing with constant per-edit overhead while explicitly controlling where edited knowledge is allowed to propagate.

Extensive experiments validate the effectiveness and generality of \lsy{ScopeEdit}. Across representative parameter-modifying and parameter-preserving editors, \lsy{ScopeEdit} achieves a more favorable edit-scoped trade-off, improving in-scope cross-modal generalization while preserving out-of-scope locality and edit reliability. Mechanistic and long-horizon analyses further confirm that its scope-separated write geometry, evidence-gated propagation, and recursive updates are essential for robust and stable online editing. Results on diverse MLLM backbones, real-world VLKEB scenarios, and more complex vision-language architectures further demonstrate that edit-scoped generalization is not tied to a specific model design, but provides a broadly applicable principle for online multimodal editing.
Future work may extend this principle toward more adaptive scope discovery, richer open-world multimodal streams, and uncertainty-aware propagation control, enabling MLLMs to support more reliable and scalable lifelong knowledge adaptation.

\bibliographystyle{IEEEtran}
\bibliography{reference}

@article{more,
  title={Modality-Decoupled Online Recursive Editing},
  author={Li, Siyuan and Zhang, Youyuan and Liu, Fangming and Li, Jing},
  journal={arXiv preprint arXiv:2605.20273},
  year={2026}
}

@inproceedings{zeng2025visual,
  title={Visual-oriented fine-grained knowledge editing for multimodal large language models},
  author={Zeng, Zhen and Gu, Leijiang and Yang, Xun and Duan, Zhangling and Shi, Zenglin and Wang, Meng},
  booktitle={Proceedings of the IEEE/CVF International Conference on Computer Vision},
  pages={2491--2500},
  year={2025}
}

@article{wang2026vlbiasbench,
  title={Vlbiasbench: A comprehensive benchmark for evaluating bias in large vision-language model},
  author={Wang, Sibo and Cao, Xiangkui and Zhang, Jie and Yuan, Zheng and Shan, Shiguang and Chen, Xilin and Gao, Wen},
  journal={IEEE Transactions on Pattern Analysis and Machine Intelligence},
  year={2026},
  publisher={IEEE}
}

@article{liu2026principled,
  title={Principled multimodal representation learning},
  author={Liu, Xiaohao and Xia, Xiaobo and Ng, See-Kiong and Chua, Tat-Seng},
  journal={IEEE Transactions on Pattern Analysis and Machine Intelligence},
  year={2026},
  publisher={IEEE}
}

@inproceedings{Huang2024vlkeb,
  author       = {Han Huang and
                  Haitian Zhong and
                  Tao Yu and
                  Qiang Liu and
                  Shu Wu and
                  Liang Wang and
                  Tieniu Tan},
  title        = {{VLKEB:} {A} Large Vision-Language Model Knowledge Editing Benchmark},
  booktitle    = {Advances in Neural Information Processing Systems 38: Annual Conference
                  on Neural Information Processing Systems 2024, NeurIPS 2024, Vancouver,
                  BC, Canada, December 10 - 15, 2024},
  year         = {2024},
}

@inproceedings{fang2025can,
  title={Can Knowledge be Transferred from Unimodal to Multimodal? Investigating the Transitivity of Multimodal Knowledge Editing},
  author={Fang, Lingyong and Wang, Xinzhong and Wang, Depeng and Wu, Zongru and Guo, Ya and Zhu, Huijia and Zhang, Zhuosheng and Liu, Gongshen},
  booktitle={Proceedings of the IEEE/CVF International Conference on Computer Vision},
  pages={2482--2490},
  year={2025}
}

@inproceedings{
jiang2026when,
title={When Large Multimodal Models Confront Evolving Knowledge: Challenges and Explorations},
author={Kailin Jiang and Yuntao Du and Yukai Ding and Yuchen Ren and Ning Jiang and Zhi Gao and Zilong Zheng and Lei Liu and Bin Li and Qing Li},
  booktitle={Proceedings of the International Conference on Learning Representations (ICLR)},
year={2026},
}

@article{lin2026moe,
  title={Moe-llava: Mixture of experts for large vision-language models},
  author={Lin, Bin and Tang, Zhenyu and Ye, Yang and Huang, Jinfa and Zhang, Junwu and Pang, Yatian and Jin, Peng and Ning, Munan and Luo, Jiebo and Yuan, Li},
  journal={IEEE Transactions on Multimedia},
  year={2026},
  publisher={IEEE}
}

@article{xue2026vision,
  title={Vision-Controllable Language Model for Image-Guided Story Ending Generation},
  author={Xue, Dizhan and Qian, Shengsheng and Xu, Changsheng},
  journal={IEEE Transactions on Multimedia},
  year={2026},
  publisher={IEEE}
}

@article{touvron2023llama,
  title={Llama: Open and efficient foundation language models},
  author={Touvron, Hugo and Lavril, Thibaut and Izacard, Gautier and Martinet, Xavier and Lachaux, Marie-Anne and Lacroix, Timoth{\'e}e and Rozi{\`e}re, Baptiste and Goyal, Naman and Hambro, Eric and Azhar, Faisal and others},
  journal={arXiv preprint arXiv:2302.13971},
  year={2023}
}

@article{zhao2023survey,
  title={A survey of large language models},
  author={Zhao, Wayne Xin and Zhou, Kun and Li, Junyi and Tang, Tianyi and Wang, Xiaolei and Hou, Yupeng and Min, Yingqian and Zhang, Beichen and Zhang, Junjie and Dong, Zican and others},
  journal={arXiv preprint arXiv:2303.18223},
  volume={1},
  number={2},
  year={2023}
}

@inproceedings{li2023blip,
  title={Blip-2: Bootstrapping language-image pre-training with frozen image encoders and large language models},
  author={Li, Junnan and Li, Dongxu and Savarese, Silvio and Hoi, Steven},
  booktitle={Proceedings of the International conference on machine learning (ICML)},
  pages={19730--19742},
  year={2023}
}

@inproceedings{lin2024video,
  title={Video-llava: Learning united visual representation by alignment before projection},
  author={Lin, Bin and Ye, Yang and Zhu, Bin and Cui, Jiaxi and Ning, Munan and Jin, Peng and Yuan, Li},
  booktitle={Proceedings of the 2024 conference on empirical methods in natural language processing (EMNLP)},
  pages={5971--5984},
  year={2024}
}

@inproceedings{cheng2023edit,
    title = "Can We Edit Multimodal Large Language Models?",
    author = "Cheng, Siyuan  and
      Tian, Bozhong  and
      Liu, Qingbin  and
      Chen, Xi  and
      Wang, Yongheng  and
      Chen, Huajun  and
      Zhang, Ningyu",
    booktitle = "Proceedings of the 2023 Conference on Empirical Methods in Natural Language Processing (EMNLP)",
    year = "2023",
    pages = "13877--13888"
}

@inproceedings{chen2025attribution,
  title={Attribution analysis meets model editing: Advancing knowledge correction in vision language models with visedit},
  author={Chen, Qizhou and Zhang, Taolin and Wang, Chengyu and He, Xiaofeng and Wang, Dakan and Liu, Tingting},
  booktitle={Proceedings of the AAAI Conference on Artificial Intelligence (AAAI)},
  pages={2168--2176},
  year={2025}
}

@inproceedings{meng2022locating,
  title={Locating and editing factual associations in gpt},
  author={Meng, Kevin and Bau, David and Andonian, Alex and Belinkov, Yonatan},
  booktitle={Proceedings of the Advances in neural information processing systems (NeurIPS)},
  pages={17359--17372},
  year={2022}
}

@inproceedings{fangalphaedit,
  title={AlphaEdit: Null-Space Constrained Knowledge Editing for Language Models},
  author={Fang, Junfeng and Jiang, Houcheng and Wang, Kun and Ma, Yunshan and Shi, Jie and Wang, Xiang and He, Xiangnan and Chua, Tat-Seng},
  booktitle={Proceedings of the International Conference on Learning Representations (ICLR)},
  year={2025}
}

@inproceedings{mitchell2022memory,
  title={Memory-based model editing at scale},
  author={Mitchell, Eric and Lin, Charles and Bosselut, Antoine and Manning, Christopher D and Finn, Chelsea},
  booktitle={Proceedings of the International Conference on Machine Learning (ICML)},
  pages={15817--15831},
  year={2022}
}

@inproceedings{
shi2025dualedit,
title={DualEdit: Dual Editing for Knowledge Updating in Vision-Language Models},
author={Zhiyi Shi and Binjie Wang and Chongjie Si and Yichen Wu and Junsik Kim and Hanspeter Pfister},
booktitle={Proceedings of the Second Conference on Language Modeling (COLM)},
  year={2025},
}

@article{cao2025deltaedit,
  title={DeltaEdit: Enhancing Sequential Editing in Large Language Models by Controlling Superimposed Noise},
  author={Cao, Ding and Cai, Yuchen and Guo, Rongxi and He, Xuesong and Liu, Guiquan},
  journal={arXiv preprint arXiv:2505.07899},
  year={2025}
}

@inproceedings{chen2025lifelong,
  title={Lifelong knowledge editing for vision language models with low-rank mixture-of-experts},
  author={Chen, Qizhou and Wang, Chengyu and Wang, Dakan and Zhang, Taolin and Li, Wangyue and He, Xiaofeng},
  booktitle={Proceedings of the Computer Vision and Pattern Recognition Conference (CVPR)},
  pages={9455--9466},
  year={2025}
}

@inproceedings{li2024can,
  title={Can we continually edit language models? on the knowledge attenuation in sequential model editing},
  author={Li, Qi and Chu, Xiaowen},
  booktitle={Findings of the Association for Computational Linguistics (ACL)},
  pages={5438--5455},
  year={2024}
}

@article{zhang2024comprehensive,
  title={A comprehensive study of knowledge editing for large language models},
  author={Zhang, Ningyu and Yao, Yunzhi and Tian, Bozhong and Wang, Peng and Deng, Shumin and Wang, Mengru and Xi, Zekun and Mao, Shengyu and Zhang, Jintian and Ni, Yuansheng and others},
  journal={arXiv preprint arXiv:2401.01286},
  year={2024}
}

@inproceedings{
meng2023massediting,
title={Mass-Editing Memory in a Transformer},
author={Kevin Meng and Arnab Sen Sharma and Alex J Andonian and Yonatan Belinkov and David Bau},
booktitle={Proceedings of the International Conference on Learning Representations (ICLR)},
year={2023},

}

@inproceedings{de-cao-etal-2021-editing,
    title = "Editing Factual Knowledge in Language Models",
    author = "De Cao, Nicola  and
      Aziz, Wilker  and
      Titov, Ivan",
    booktitle = "Proceedings of the 2021 Conference on Empirical Methods in Natural Language Processing (EMNLP)",
    year = "2021"
}

@inproceedings{DBLP:conf/iclr/MitchellLBFM22,
  author       = {Eric Mitchell and
                  Charles Lin and
                  Antoine Bosselut and
                  Chelsea Finn and
                  Christopher D. Manning},
  title        = {Fast Model Editing at Scale},
  booktitle    = {Proceedings of the International Conference on Learning Representations (ICLR)},
  year         = {2022}
}

@inproceedings{DBLP:conf/nips/HartvigsenSPKG23,
  author       = {Tom Hartvigsen and
                  Swami Sankaranarayanan and
                  Hamid Palangi and
                  Yoon Kim and
                  Marzyeh Ghassemi},
  title        = {Aging with {GRACE:} Lifelong Model Editing with Discrete Key-Value
                  Adaptors},
  booktitle    = {Proceedings of the Advances in neural information processing systems (NeurIPS)},
  year         = {2023}
}

@inproceedings{geva2021transformer,
  title={Transformer feed-forward layers are key-value memories},
  author={Geva, Mor and Schuster, Roei and Berant, Jonathan and Levy, Omer},
  booktitle={Proceedings of the 2021 Conference on Empirical Methods in Natural Language Processing (EMNLP)},
  pages={5484--5495},
  year={2021}
}

@inproceedings{dai-etal-2022-knowledge,
    title = "Knowledge Neurons in Pretrained Transformers",
    author = "Dai, Damai  and
      Dong, Li  and
      Hao, Yaru  and
      Sui, Zhifang  and
      Chang, Baobao  and
      Wei, Furu",
    booktitle = "Proceedings of the 60th Annual Meeting of the Association for Computational Linguistics (ACL)",
    year = "2022",
    pages = "8493--8502"
}

@inproceedings{DBLP:conf/iclr/SinitsinPPPB20,
  author       = {Anton Sinitsin and
                  Vsevolod Plokhotnyuk and
                  Dmitry V. Pyrkin and
                  Sergei Popov and
                  Artem Babenko},
  title        = {Editable Neural Networks},
  booktitle    = {Proceedings of the International Conference on Learning Representations (ICLR)},
  year         = {2020}
}

@inproceedings{DBLP:conf/iclr/HuangSZZR023,
  author       = {Zeyu Huang and
                  Yikang Shen and
                  Xiaofeng Zhang and
                  Jie Zhou and
                  Wenge Rong and
                  Zhang Xiong},
  title        = {Transformer-Patcher: One Mistake Worth One Neuron},
  booktitle    = {Proceedings of the International Conference on Learning Representations (ICLR)},
  year         = {2023}
}

@inproceedings{DBLP:conf/aaai/YuCZH24,
  author       = {Lang Yu and
                  Qin Chen and
                  Jie Zhou and
                  Liang He},
  title        = {{MELO:} Enhancing Model Editing with Neuron-Indexed Dynamic LoRA},
  booktitle    = {Proceedings of the AAAI Conference on Artificial Intelligence (AAAI)},
  pages        = {19449--19457},
  year         = {2024}
}

@article{he2024llmsmeetmultimodalgeneration,
  title={LLMs Meet Multimodal Generation and Editing: A Survey},
  author={Yingqing He and Zhaoyang Liu and Jingye Chen and Zeyue Tian and Hongyu Liu and Xiaowei Chi and Runtao Liu and Ruibin Yuan and Yazhou Xing and Wenhai Wang and Jifeng Dai and Yong Zhang and Wei Xue and Qifeng Liu and Yike Guo and Qifeng Chen},
  journal={arXiv preprint arXiv:2405.19334},
  year={2024}
}

@inproceedings{jiang-etal-2025-neuron,
    title = "Neuron-Level Sequential Editing for Large Language Models",
    author = "Jiang, Houcheng  and
      Fang, Junfeng  and
      Zhang, Tianyu  and
      Bi, Baolong  and
      Zhang, An  and
      Wang, Ruipeng  and
      Liang, Tao  and
      Wang, Xiang",
    booktitle = "Proceedings of the 63rd Annual Meeting of the Association for Computational Linguistics (ACL)",
    year = "2025",
    pages = "16678--16702"
}

@inproceedings{DBLP:conf/nips/0104L0XY0X0C24,
  author       = {Peng Wang and
                  Zexi Li and
                  Ningyu Zhang and
                  Ziwen Xu and
                  Yunzhi Yao and
                  Yong Jiang and
                  Pengjun Xie and
                  Fei Huang and
                  Huajun Chen},
  title        = {{WISE:} Rethinking the Knowledge Memory for Lifelong Model Editing
                  of Large Language Models},
  booktitle    = {Proceedings of the Advances in Neural Information Processing Systems (NeurIPS)},
  year         = {2024}
}

@inproceedings{chen-etal-2024-lifelong,
    title = "Lifelong Knowledge Editing for {LLM}s with Retrieval-Augmented Continuous Prompt Learning",
    author = "Chen, Qizhou  and
      Zhang, Taolin  and
      He, Xiaofeng  and
      Li, Dongyang  and
      Wang, Chengyu  and
      Huang, Longtao  and
      Xue{'}, Hui",
    booktitle = "Proceedings of the 2024 Conference on Empirical Methods in Natural Language Processing (EMNLP)",
    year = "2024",
    pages = "13565--13580"
}

@inproceedings{DBLP:conf/icml/ThedeRBAH25,
  author       = {Lukas Thede and
                  Karsten Roth and
                  Matthias Bethge and
                  Zeynep Akata and
                  Thomas Hartvigsen},
  title        = {WikiBigEdit: Understanding the Limits of Lifelong Knowledge Editing
                  in LLMs},
  booktitle    = {Proceedings of the International Conference on Machine Learning (ICML)},
  year         = {2025}
}

@inproceedings{DBLP:conf/emnlp/ZhengLDFWXC23,
  author       = {Ce Zheng and
                  Lei Li and
                  Qingxiu Dong and
                  Yuxuan Fan and
                  Zhiyong Wu and
                  Jingjing Xu and
                  Baobao Chang},
  title        = {Can We Edit Factual Knowledge by In-Context Learning?},
  booktitle    = {Proceedings of the 2023 Conference on Empirical Methods in Natural
                  Language Processing (EMNLP)},
  pages        = {4862--4876},
  year         = {2023}
}

@inproceedings{DBLP:conf/cvpr/LiuLLL24,
  author       = {Haotian Liu and
                  Chunyuan Li and
                  Yuheng Li and
                  Yong Jae Lee},
  title        = {Improved Baselines with Visual Instruction Tuning},
  booktitle    = {Proceedings of the Computer Vision and Pattern Recognition Conference (CVPR)},
  pages        = {26286--26296},
  year         = {2024}
}

@article{wang2024qwen2vlenhancingvisionlanguagemodels,
  title={Qwen2-VL: Enhancing Vision-Language Model's Perception of the World at Any Resolution},
  author={Peng Wang and Shuai Bai and Sinan Tan and Shijie Wang and Zhihao Fan and Jinze Bai and Keqin Chen and Xuejing Liu and Jialin Wang and Wenbin Ge and Yang Fan and Kai Dang and Mengfei Du and Xuancheng Ren and Rui Men and Dayiheng Liu and Chang Zhou and Jingren Zhou and Junyang Lin},
  journal={arXiv preprint arXiv:2409.12191},
  year={2024}
}

@article{bai2025qwen3vltechnicalreport,
  title={Qwen3-VL Technical Report},
  author={Shuai Bai and Yuxuan Cai and Ruizhe Chen and Keqin Chen and Xionghui Chen and Zesen Cheng and Lianghao Deng and Wei Ding and Chang Gao and Chunjiang Ge and Wenbin Ge and Zhifang Guo and Qidong Huang and Jie Huang and Fei Huang and Binyuan Hui and Shutong Jiang and Zhaohai Li and Mingsheng Li and Mei Li and Kaixin Li and Zicheng Lin and Junyang Lin and Xuejing Liu and Jiawei Liu and Chenglong Liu and Yang Liu and Dayiheng Liu and Shixuan Liu and Dunjie Lu and Ruilin Luo and Chenxu Lv and Rui Men and Lingchen Meng and Xuancheng Ren and Xingzhang Ren and Sibo Song and Yuchong Sun and Jun Tang and Jianhong Tu and Jianqiang Wan and Peng Wang and Pengfei Wang and Qiuyue Wang and Yuxuan Wang and Tianbao Xie and Yiheng Xu and Haiyang Xu and Jin Xu and Zhibo Yang and Mingkun Yang and Jianxin Yang and An Yang and Bowen Yu and Fei Zhang and Hang Zhang and Xi Zhang and Bo Zheng and Humen Zhong and Jingren Zhou and Fan Zhou and Jing Zhou and Yuanzhi Zhu and Ke Zhu},
  journal={arXiv preprint arXiv:2511.21631},
  year={2025}
}

@inproceedings{du2025mmke,
    title={MMKE-Bench: A Multimodal Editing Benchmark for Diverse Visual Knowledge},
    author={Du, Yuntao and Jiang, Kailin and Gao, Zhi and Shi, Chenrui and Zheng, Zilong and Qi, Siyuan and Li, Qing},
    booktitle={The Thirteenth International Conference on Learning Representations(ICLR)},
    year={2025}
}

@inproceedings{zhang-etal-2025-mc,
    title = "{MC}-{MKE}: A Fine-Grained Multimodal Knowledge Editing Benchmark Emphasizing Modality Consistency",
    author = "Zhang, Junzhe  and
      Zhang, Huixuan  and
      Yin, Xunjian  and
      Huang, Baizhou  and
      Zhang, Xu  and
      Hu, Xinyu  and
      Wan, Xiaojun",
    booktitle = "Findings of the Association for Computational Linguistics(ACL)",
    year = "2025",
    pages = "17430--17445"
}

@inproceedings{zhou-etal-2025-m2edit,
    title = "{M}2{E}dit: Locate and Edit Multi-Granularity Knowledge in Multimodal Large Language Model",
    author = "Zhou, Yang  and
      Cao, Pengfei  and
      Chen, Yubo  and
      Liu, Qingbin  and
      Sui, Dianbo  and
      Chen, Xi  and
      Liu, Kang  and
      Zhao, Jun",
    booktitle = "Proceedings of the 2025 Conference on Empirical Methods in Natural Language Processing(EMNLP)",
    year = "2025",
    pages = "29029--29042"
}

@InProceedings{Das_2026_CVPR,
    author    = {Das, Gyanendra and Jena, Sai},
    title     = {DSCA: Dynamic Subspace Concept Alignment for Lifelong VLM Editing},
    booktitle = {Proceedings of the IEEE/CVF Conference on Computer Vision and Pattern Recognition (CVPR)},
    year      = {2026},
    pages     = {40772-40781}
}

@InProceedings{pmlr-v267-guo25e,
  title = 	 {{B}alanc{E}dit: Dynamically Balancing the Generality-Locality Trade-off in Multi-modal Model Editing},
  author =       {Guo, Dongliang and Hu, Mengxuan and Guan, Zihan and Hartvigsen, Thomas and Li, Sheng},
  booktitle = 	 {Proceedings of the 42nd International Conference on Machine Learning(ICML)},
  pages = 	 {20843--20857},
  year = 	 {2025},
  volume = 	 {267}
}

@inproceedings{NEURIPS2025_bbdd7f4a,
 author = {Seong, Jin and Park, Jiyun and Liermann, Wencke and Choi, Hongseok and Nam, Yoonji and Kim, Hyun and Lim, Soojong and Lee, Namhoon},
 booktitle = {Proceedings of the Advances in neural information processing systems (NeurIPS)},
 pages = {129205--129242},
 title = {MemEIC: A Step Toward Continual and Compositional Knowledge Editing},
 volume = {38},
 year = {2025}
}

@inproceedings{jia-etal-2025-exploring,
    title = "Exploring and Evaluating Multimodal Knowledge Reasoning Consistency of Multimodal Large Language Models",
    author = "Jia, Boyu  and
      Zhang, Junzhe  and
      Zhang, Huixuan  and
      Wan, Xiaojun",
    booktitle = "Findings of the Association for Computational Linguistics(EMNLP)",
    year = "2025",
    pages = "11966--11981"
}

@inproceedings{wang-etal-2025-v,
    title = "{V}-{SEAM}: Visual Semantic Editing and Attention Modulating for Causal Interpretability of Vision-Language Models",
    author = "Wang, Qidong  and
      Hu, Junjie  and
      Jiang, Ming",
    booktitle = "Proceedings of the 2025 Conference on Empirical Methods in Natural Language Processing(EMNLP)",
    year = "2025",
    pages = "17396--17420"
}

@inproceedings{liu-etal-2025-insight,
    title = "Insight Over Sight: Exploring the Vision-Knowledge Conflicts in Multimodal {LLM}s",
    author = "Liu, Xiaoyuan  and
      Wang, Wenxuan  and
      Yuan, Youliang  and
      Huang, Jen-tse  and
      Liu, Qiuzhi  and
      He, Pinjia  and
      Tu, Zhaopeng",
    booktitle = "Proceedings of the 63rd Annual Meeting of the Association for Computational Linguistics (ACL)",
    year = "2025",
    pages = "17825--17846"
}

@inproceedings{shao-etal-2025-cognition,
    title = "Is Cognition Consistent with Perception? Assessing and Mitigating Multimodal Knowledge Conflicts in Document Understanding",
    author = "Shao, Zirui  and
      Gao, Feiyu  and
      Zhu, Zhaoqing  and
      Luo, Chuwei  and
      Xing, Hangdi  and
      Yu, Zhi  and
      Zheng, Qi  and
      Yan, Ming  and
      Bu, Jiajun",
    booktitle = "Proceedings of the 2025 Conference on Empirical Methods in Natural Language Processing(EMNLP)",
    year = "2025",
    pages = "30923--30944"
}

@InProceedings{pmlr-v235-ma24h,
  title = 	 {Neighboring Perturbations of Knowledge Editing on Large Language Models},
  author =       {Ma, Jun-Yu and Ling, Zhen-Hua and Zhang, Ningyu and Gu, Jia-Chen},
  booktitle = 	 {Proceedings of the 41st International Conference on Machine Learning(ICML)},
  pages = 	 {33839--33854},
  year = 	 {2024},
  volume = 	 {235}
}

@inproceedings{DBLP:conf/iclr/ZhangYLWRC25,
  author       = {Mengqi Zhang and
                  Xiaotian Ye and
                  Qiang Liu and
                  Shu Wu and
                  Pengjie Ren and
                  Zhumin Chen},
  title        = {Uncovering Overfitting in Large Language Model Editing},
  booktitle    = {Proceedings of the Thirteenth International Conference on Learning Representations(ICLR)},
  year         = {2025}
}

@inproceedings{ICLR2025_35cb54b8,
 author = {Wang, Pinzheng and Tang, Zecheng and Zhou, Keyan and Li, Juntao and Zhu, Qiaoming and Zhang, Min},
  booktitle    = {Proceedings of the Thirteenth International Conference on Learning Representations(ICLR)},
 pages = {20922--20948},
 title = {Revealing and Mitigating Over-Attention in Knowledge Editing},
 year = {2025}
}

@InProceedings{pmlr-v235-wang24s,
  title = 	 {{MEMORYLLM}: Towards Self-Updatable Large Language Models},
  author =       {Wang, Yu and Gao, Yifan and Chen, Xiusi and Jiang, Haoming and Li, Shiyang and Yang, Jingfeng and Yin, Qingyu and Li, Zheng and Li, Xian and Yin, Bing and Shang, Jingbo and Mcauley, Julian},
  booktitle = 	 {Proceedings of the 41st International Conference on Machine Learning(ICML)},
  pages = 	 {50453--50466},
  year = 	 {2024}
}

@inproceedings{ICLR2025_2f89a23a,
 author = {Liu, Tianci and Li, Ruirui and Qi, Yunzhe and Liu, Hui and Tang, Xianfeng and Zheng, Tianqi and Yin, Qingyu and Cheng, Monica and Huan, Jun and Wang, Haoyu and Gao, Jing},
  booktitle    = {Proceedings of the Thirteenth International Conference on Learning Representations(ICLR)},  
 pages = {18939--18959},
 title = {Unlocking Efficient, Scalable, and Continual Knowledge Editing with Basis-Level Representation Fine-Tuning},
 year = {2025}
}

@inproceedings{3737916.3738887,
author = {Li, Qi and Liu, Xiang and Tang, Zhenheng and Dong, Peijie and Li, Zeyu and Pan, Xinglin and Chu, Xiaowen},
title = {Should we really edit language models? on the evaluation of edited language models},
year = {2024},
booktitle = {Proceedings of the 38th International Conference on Neural Information Processing Systems(NeurIPS)}
}

@inproceedings{zhai-etal-2025-parameter,
    title = "Parameter-Aware Contrastive Knowledge Editing: Tracing and Rectifying based on Critical Transmission Paths",
    author = "Zhai, Songlin  and
      Meng, Yuan  and
      Zhang, Yuxin  and
      Qi, Guilin",
    booktitle = "Proceedings of the 63rd Annual Meeting of the Association for Computational Linguistics (ACL)",
    year = "2025",
    pages = "28189--28200"
}

@inproceedings{yao-etal-2025-cake,
    title = "{C}a{KE}: Circuit-aware Editing Enables Generalizable Knowledge Learners",
    author = "Yao, Yunzhi  and
      Fang, Jizhan  and
      Gu, Jia-Chen  and
      Zhang, Ningyu  and
      Deng, Shumin  and
      Chen, Huajun  and
      Peng, Nanyun",
    booktitle = "Proceedings of the 2025 Conference on Empirical Methods in Natural Language Processing(EMNLP)",
    year = "2025",
    pages = "11366--11382"
}

@inproceedings{scialanga-etal-2025-sake,
    title = "{SAKE}: Steering Activations for Knowledge Editing",
    author = "Scialanga, Marco  and
      Laugel, Thibault  and
      Grari, Vincent  and
      Detyniecki, Marcin",
    booktitle = "Proceedings of the 63rd Annual Meeting of the Association for Computational Linguistics (ACL)",
    year = "2025",
    pages = "15966--15978"
}

@inproceedings{wang-etal-2025-knowledge-editing,
    title = "Knowledge Editing through Chain-of-Thought",
    author = "Wang, Changyue  and
      Su, Weihang  and
      Ai, Qingyao  and
      Tang, Yichen  and
      Liu, Yiqun",
    booktitle = "Proceedings of the 2025 Conference on Empirical Methods in Natural Language Processing(EMNLP)",
    year = "2025",
    pages = "10673--10693"
}

@inproceedings{li-chu-2025-adaedit,
    title = "{A}da{E}dit: Advancing Continuous Knowledge Editing For Large Language Models",
    author = "Li, Qi  and
      Chu, Xiaowen",
    booktitle = "Proceedings of the 63rd Annual Meeting of the Association for Computational Linguistics (ACL)",
    year = "2025",
    pages = "4127--4149"
}

\end{document}